\documentclass{article}
\usepackage{graphicx}
\graphicspath{{plots/}}
\DeclareGraphicsExtensions{.pdf,.png,.jpg}
\usepackage[T1]{fontenc}
\usepackage[utf8]{inputenc}

\usepackage{microtype}
\usepackage{graphicx}
\usepackage{subfigure}
\usepackage{booktabs} 

\usepackage{algorithm}
\usepackage{algorithmic}
\usepackage{amsmath}
\usepackage{amsfonts}
\usepackage{amsthm}
\usepackage{mathtools}
\usepackage{verbatim}
\usepackage{hyperref}
\usepackage{natbib}

\newtheorem{theorem}{Theorem}[section]
\newtheorem{corollary}[theorem]{Corollary}
\newtheorem{lemma}[theorem]{Lemma}
\newtheorem{assumption}[theorem]{Assumption}
\newtheorem{remark}[theorem]{Remark}

\usepackage[colorinlistoftodos,bordercolor=orange,backgroundcolor=orange!20,linecolor=orange,textsize=scriptsize]{todonotes}

\newcommand{\xx}{\hat{x}^0}

\usepackage[accepted]{icml_modified}

\icmltitlerunning{From Local SGD to Local Fixed-Point Methods for Federated Learning}

\begin{document}

	\twocolumn[
	\icmltitle{From Local SGD to Local Fixed-Point Methods for Federated Learning}



	\icmlsetsymbol{equal}{*}

	\begin{icmlauthorlist}
		\icmlauthor{Grigory Malinovsky}{goo}
		\icmlauthor{Dmitry Kovalev}{to}
		\icmlauthor{Elnur Gasanov}{to}
		\icmlauthor{Laurent Condat}{to}
		\icmlauthor{Peter Richt\'arik}{to}
			\end{icmlauthorlist}

	\icmlaffiliation{to}{King Abdullah University of Science and Technology (KAUST), Thuwal, Saudi Arabia}
	\icmlaffiliation{goo}{Moscow Institute of Physics and Technology}

	\icmlcorrespondingauthor{Laurent Condat}{see https://lcondat.github.io/}

	\icmlkeywords{Machine Learning, ICML}

	\vskip 0.3in
	]



	\printAffiliationsAndNotice{}  

	\begin{abstract}
Most algorithms for solving optimization problems or finding saddle points of convex--concave functions are fixed-point algorithms. In this work we consider the generic problem of finding a fixed point of an average of operators, or an approximation thereof, in a distributed setting. Our work is motivated by the needs of federated learning. In this context, each local operator models the computations done locally on a mobile device. We investigate two strategies to achieve such a consensus: one  based on a fixed number of  local steps, and the other based on randomized computations. In both cases, the goal is to limit communication of the locally-computed variables, which is often the bottleneck in distributed frameworks. We perform convergence analysis of both methods and conduct a number of experiments highlighting the benefits of our approach.
	\end{abstract}

	\section{Introduction}

	In the `big data' era, the explosion in size and complexity of the data
	arises in parallel to a shift towards distributed computations, as modern hardware increasingly relies on the power of uniting many parallel units into one system. For distributed optimization tasks, specific issues arise, such as
	decentralized data storage.  For instance, the huge amount of mobile phones or  smart home devices in the world contain an important volume of data  captured and stored on each of them. This data contains a wealth of potentially useful information to their owners, and more so if appropriate machine learning models could be trained on the heterogeneous data stored across the network of such devices.  Yet, many users are increasingly sensitive to privacy concerns and prefer their data to never leave their devices. But the only way to share knowledge while not having all data in one place is to communicate, to keep moving towards the solution of the overall problem. Typically, mobile phones communicate back and forth with a distant server, 
	so that a global model is progressively improved and converges
	to a steady state, which is globally optimal  for all users.
	This is precisely the purpose of the recent and rising paradigm of \emph{federated learning}
	\cite{ja2016, FL2017-AISTATS}
	where typically a global supervised model is trained in a massively distributed manner over a network of \emph{heterogeneous} devices.
	Communication, which can be costly and slow, is the main bottleneck in this framework. So, it is of primary importance to devise novel algorithmic strategies, where the computation and communication loads are balanced.


	A strategy increasingly used by practitioners is to make use of \emph{local computations};
	that is, more local computations are performed on each device before communication and subsequent model averaging, with the hope  that this will reduce the total number of communications needed to obtain a globally meaningful solution. Thus, local gradient descent methods have been investigated~\cite{Stich2018,localGD,localSGD-AISTATS2020,ma2017distributed,haddadpour2019convergence}.
	Despite their practical success, local methods are little understood and there is much to be discovered. In this paper, we don't restrict ourselves to gradient descent to minimize an average of smooth functions; we consider the much broader setting of finding a fixed point of an average of a large number $M$ of operators.
	%
	%
	Indeed, most, if not all, iterative methods are fixed-point methods, which aim at finding a fixed point of some operator \cite{bau11a}. Fixed-point methods are typically made from compositions and averages of gradient or proximity operators of functions \cite{comy15,bau17}; for instance, a sum of proximity operators corresponds to the `proximal average' of functions \cite{yu13}. Using more involved Lyapunov functions than the distance to the solution or the objective value, convergence of methods with inertia, e.g. Nesterov's acceleration techniques, to a fixed point, can be established \cite{les16}. 
(Block-)coordinate or alternating minimization methods are fixed-point methods as well \cite{ric14,pes15}. Let us also mention that by the design of nontrivial fixed-point operators, nonlinear inverse problems can be solved \cite{com20}. Beyond optimization, fixed-point methods are used to solve monotone inclusions or variational inequalities, with applications in mechanics or stochastic control. They are also  used to find saddle points of convex--concave functions, e.g.\ Nash equilibria in game theory. Yet another example is attaining the steady-state of a control system or a dynamic phenomenon modeled by a PDE.

	%

	\subsection{Contributions}
	We model the setting of communication-efficient distributed fixed-point optimization as follows:
	we have $M\geq 1$ parallel computing nodes. The variables handled by these nodes are modeled as vectors of the Euclidean space $\mathbb{R}^d$, endowed with the classical inner product, for some $d\geq 1$.
	Let $\mathcal{T}_i$, for $i=1,\ldots,M$ be operators on $\mathbb{R}^d$, which model the set of operations during one iteration. We define the average operator
	\begin{equation}
	\label{T}
	\mathcal{T}:x\in\mathbb{R}^d\mapsto \frac{1}{M} \sum_{i=1}^{M} \mathcal{T}_{i}(x).
	\end{equation}
	Our goal is to find  a fixed point of $\mathcal{T}$; that is, a vector $x^\star\in\mathbb{R}^d$ such that
	$\mathcal{T}(x^\star) = x^\star$.
	The sought solution $x^\star$ should be obtained by repeatedly applying $\mathcal{T}_i$ at each node, in parallel, with averaging steps to achieve a consensus. Here we consider that, after some number of iterations, each node communicates its variable to a distant server, synchronously. Then the server computes the average of the received vectors and broadcasts it to all nodes.


We investigate two strategies. The first one consists,
  for each computing node, in iterating several times some sequence of operations; we call this \emph{local steps}. 
The second strategy consists in reducing
  the number of communication steps by sharing information only with some low probability, and doing only local computations inbetween. We analyze two algorithms, which instantiate these two ideas, and we prove their convergence. Their good performances are illustrated by experiments.

	\subsection{Mathematical Background}

	Let $T$ be an operator on $\mathbb{R}^d$. We denote by $\mathrm{Fix}(T)$ the set of its fixed points.
$T$ is said to be $\chi$-Lipschitz continuous, for some $\chi\geq 0$, if, for every $x$ and $y$ in $\mathbb{R}^d$,
	\begin{equation*}
	\|T(x)-T(y)\|\leq \chi \|x-y\|.
	\end{equation*}
	Moreover, $T$ is said to be nonexpansive if it is 1-Lipschitz continuous, and $\chi$-contractive, if it is $\chi$-Lipschitz continuous, for some $\chi \in [0,1)$.
	If $T$ is contractive, its fixed point exists and is unique, see the Banach--Picard Theorem 1.50 in \cite{bau17}.
$T$ is said to be $\alpha$-averaged, for some $\alpha\in(0,1]$, if $T=\alpha T'+\mathrm{Id}$ for some nonexpansive operator $T'$, where $\mathrm{Id}$ denotes the identity. $T$ is said to be firmly nonexpansive if it is $1/2$-averaged.

	\section{A Generic Distributed Fixed-Point Method with Local Steps}

Let $(t_n)_{n\in\mathbb{N}}$ be the sequence of integers at which communication occurs. We propose Algorithm 1, shown below; it proceeds as follows: at every iteration, the operator $\mathcal{T}_i$ is applied at node $i$, with further relaxation with parameter $\lambda$. After some number of iterations, each of the $M$ computing nodes communicates its vector to a master node, which computes their average and broadcasts it to all nodes. Thus, the later resume computing at the next iteration from the same variable $\hat{x}^k$. The algorithm is a generalization of local gradient descent, a.k.a. Federated Averaging \cite{FL2017-AISTATS}.
	

We call an \emph{epoch} a sequence of local iterations, followed by averaging; that is, the $n$-th epoch, for $n\geq 1$, is the sequence of iterations of indices $k+1=t_{n-1}+1,\ldots,t_{n}$ (the 0-th epoch is the initialization step $x_i^0 \coloneqq \xx$, for $i =1,\ldots,M$).
	We assume that the number of iterations in each epoch, between two aggregation steps, is bounded by some integer $H\geq 1$; that is,

	{\assumption\label{ass:H} $1\leq t_{n}-t_{n-1} \leq H$, for every $n\geq 1$.}



\begin{figure*}[t]
\begin{minipage}{.48\textwidth}
	\begin{algorithm}[H]
		\caption{Local  fixed-point method}
		\label{alg}
		\begin{algorithmic}
			\STATE
			\noindent \textbf{Input:} Initial estimate $\xx \in \mathbb{R}^d$, stepsize 

\noindent$\lambda>0$,
			sequence of synchronization times 
			
			\noindent$0=t_0<t_1<\ldots$
			\STATE \textbf{Initialize:} $x_i^0 \coloneqq \xx$, for all $i =1,\ldots,M$
			\FOR{$k=0, 1, \ldots$}
			\FOR{$i=1, 2, \ldots, M$ in parallel}
			\STATE $h_i^{k+1}\coloneqq(1-\lambda)x^{k}_{i}+\lambda \mathcal{T}_i(x^k_i)$
			\IF{$k+1 = t_n$, for some $n\in\mathbb{N}$,}
			\STATE Communicate $h_i^{k+1}$ to master node
			\ELSE
			\STATE $x^{k+1}_{i}\coloneqq h_i^{k+1}$
			\ENDIF
			\ENDFOR
			\IF{$k+1 = t_n$, for some $n\in\mathbb{N}$,}
			\STATE At master node: $\hat{x}^{k+1}\coloneqq\frac{1}{M}\sum_{i=1}^M h_i^{k+1}$%
			\STATE Broadcast: $x^{k+1}_{i}\coloneqq\hat{x}^{k+1}$, for all  $i$
			\ENDIF
			\ENDFOR
		\end{algorithmic}
	\end{algorithm}
	\end{minipage}
	\ \ \ \ \ \ \begin{minipage}{.48\textwidth}
	\begin{algorithm}[H]
		\caption{Randomized fixed-point method}
		\label{alg2}
		\begin{algorithmic}
			\STATE
			\noindent \textbf{Input:} Initial estimate $\xx \in \mathbb{R}^d$, stepsize 
			
			\noindent $\lambda>0$, communication probability $0< p\leq 1$%
\STATE \textbf{Initialize:} $x_i^0 = \xx$, for all $i =1,\ldots,M$
			\FOR{$k=1, 2, \ldots$}
			\FOR{$i=1, 2, \ldots, M$ in parallel}
			\STATE $h_i^{k+1}\coloneqq(1-\lambda)x^{k}_{i}+\lambda \mathcal{T}_i( x^k_i)$
			\ENDFOR
			\STATE Flip a coin and
			\STATE \textbf{with probability} $p$ \textbf{do}
			\STATE \ \ \ \ Communicate $h_i^{k+1}$ to master, for all $i$
			\STATE \ \ \ \ At master node: $\hat{x}^{k+1}\coloneqq\frac{1}{M}\sum_{i=1}^M h_i^{k+1}$%
			\STATE \ \ \ \ Broadcast: $x^{k+1}_{i}\coloneqq\hat{x}^{k+1}$, for all $i$ 
			\STATE \textbf{else, with probability} $1-p$, \textbf{do}
			\STATE \ \ \ \ $x^{k+1}_{i}\coloneqq h_i^{k+1}$, for all $i =1,\ldots,M$
			\ENDFOR 
		\end{algorithmic}
	\end{algorithm}\ 
	\end{minipage}
	\end{figure*}


	To analyze Algorithm 1, we introduce the following averaged vector:
	\begin{equation*}
	\hat{x}^{k} = \frac{1}{M} \sum_{i=1}^{M} x_{i}^{k}= \frac{1}{M} \sum_{i=1}^{M} h_{i}^{k}.
	\end{equation*}
	Note that this vector is actually computed only when $k$ is one of the $t_n$. 	In the uniform case $t_n=nH$, for every $n\in\mathbb{N}$, we introduce the operator
	\begin{equation*}
	\widetilde{\mathcal{T}}_{\lambda}=\frac{1}{M}\sum_{i=1}^M  \big(\lambda \mathcal{T}_i + (1-\lambda)\mathrm{Id}\big)^H,
	\end{equation*}
	where $\cdot^H$ denotes the composition of an operator with itself $H$ times.
	Thus, $x_1^{nH}=\cdots = x_M^{nH}=\hat{x}^{nH}$ is the variable shared by every node at the end of  the $n$-th epoch. We have, for every $n\in \mathbb{N}$,
	\begin{equation*}
	\hat{x}^{(n+1)H}=\widetilde{\mathcal{T}}_{\lambda}( \hat{x}^{nH}).
	\end{equation*}


	We also assume that the following holds:\medskip

	{\assumption\label{ass:fixpointexist} $\mathrm{Fix}(\mathcal{T})$ and $\mathrm{Fix}(\widetilde{\mathcal{T}}_\lambda)$ are nonempty.}

	Note that the fixed points of $\widetilde{\mathcal{T}}_\lambda$ depend on $\lambda$. The smaller $\lambda$, the closer $\mathrm{Fix}(\mathcal{T})$ and $\mathrm{Fix}(\widetilde{\mathcal{T}})$. But the smaller $\lambda$, the slower the convergence, so $\lambda$ controls the tradeoff between accuracy and speed in estimating a fixed point of $\mathcal{T}$.

\subsection{General convergence analysis}


	
	\begin{theorem}[\textbf{General convergence}]\label{thh1}
			 Suppose that $t_n=nH$, for every $n\in\mathbb{N}$, and suppose that the $\mathcal{T}_i$ are all $\alpha$-averaged, for some $\alpha\in (0,1]$. Let $\lambda\in (0,1/\alpha)$ be the parameter in Algorithm 1.
		Then the sequence $(\hat{x}^{nH})_{n\in\mathbb{N}}$ converges to a fixed point $x^\dagger$ of $\widetilde{\mathcal{T}}$. In addition, the following hold:
		
		\noindent(i) $\widetilde{\mathcal{T}}_{\lambda}$ is $\zeta$-averaged, with
		$
		\zeta = \frac{H\alpha\lambda}{1+(H-1)\alpha\lambda}.
		$
		
		\noindent(ii) The distance between $\hat{x}^{nH}$ and $x^\dagger$ decreases at every epoch: for every $n\in\mathbb{N}$,
		\begin{equation}
		\|\hat{x}^{(n+1)H}-x^\dagger\|^2\leq \|\hat{x}^{nH}-x^\dagger\|^2-\frac{1-\zeta}{\zeta}\|\hat{x}^{(n+1)H}-\hat{x}^{nH}\|^2.\label{eq011}
		\end{equation}
		
		\noindent(iii) The squared differences between two successive updates are summable:
		\begin{equation}
		\sum_{n\in\mathbb{N}}\|\hat{x}^{(n+1)H}-\hat{x}^{nH}\|^2\leq \frac{\zeta}{1-\zeta}\|\hat{x}^{0}-x^\dagger\|^2.
		\end{equation}
\noindent(iv) For every $n\in\mathbb{N}$,
\begin{equation}
		\|\hat{x}^{(n+1)H}-\hat{x}^{nH}\|^2\leq \frac{1}{\zeta(1-\zeta)(n+1)}\|\hat{x}^0-x^\dagger\|^2.
\end{equation}
\begin{equation}
	\hspace{-21mm}\mbox{(v)}\qquad\qquad 	\|\hat{x}^{(n+1)H}-\hat{x}^{nH}\|^2=o(1/n).
\end{equation}
	\end{theorem}	
	\noindent\textit{Proof}. The convergence property and the property $(iii)$ come from the application of the Krasnosel'skii--Mann theorem, see Theorem 5.15 in \cite{bau17}. The properties \emph{(i)} and \emph{(ii)} are applications of Proposition 4.46, Proposition 4.42, and Proposition 4.35 in  \cite{bau17}. \emph{(iv)} and \emph{(v)} come from Theorem~1 in  \cite{dav16}.\hfill$\square$ 
	
We can note that in most cases, the fixed-point residual $\|T(x^{k})-x^k\|$ is a natural way to measure the convergence speed of a  fixed-point algorithm $x^{k+1}=T(x^{k})$. For gradient descent, $T(x^{k})=x^k-\gamma \nabla F(x^k)$, so we have $\|T(x^{k})-x^k\|=\gamma \|\nabla F(x^k)\|$, which indeed measures the discrepancy to $\nabla F(x^\star)=0$. For the proximal point algorithm to solve a monotone inclusion $0\in M(x^\star)$, $T(x^{k})=(\gamma M +\mathrm{Id})^{-1} (x^k)$, so that $T(x^{k})-x^k\in -\gamma M(x^{k+1})$; again, $\|T(x^{k})-x^k\|$ characterizes the discrepancy to the solution.

	\begin{remark}[\textbf{Convergence speed}] For the baseline algorithm (Algorithm 1 with $H=1$), where averaging occurs after every iteration, we have after $H$ iterations:
		\begin{align}
		\|\hat{x}^{(n+1)H}-x^\dagger\|^2&\leq \|\hat{x}^{nH}-x^\dagger\|^2\label{eq012}\\
		&\quad-\frac{1-\alpha\lambda}{\alpha\lambda}\sum_{k=nH}^{(n+1)H-1}\|\hat{x}^{k+1}-\hat{x}^{k}\|^2.\notag
		\end{align}
		We can compare this `progress', made in decreasing the squared distance to the solution, with the one in Theorem \ref{thh1}-(ii), where $\frac{1-\zeta}{\zeta}=\frac{1-\alpha\lambda}{H\alpha\lambda}$. This latter value multiplies $\|\hat{x}^{(n+1)H}-\hat{x}^{nH}\|^2$, which can be up to $H^2$ larger than $\|\hat{x}^{k+1}-\hat{x}^{k}\|^2$, for $k$ in $nH,\ldots, (n+1)H-1$. So, in favorable cases, Algorithm 1 progresses as fast as the baseline algorithm. In less favorable cases, the progress in one epoch is $H$ times smaller, corresponding to the progress in 1 iteration. Given that communication occurs only once per epoch, the ratio of convergence speed to communication burden is, roughly speaking, between 1 and $H$ times better than the one of the baseline algorithm. They don't converge to the same elements, however.
		\end{remark}

			A complementary result on the convergence speed
	is the following. In the rest of the section, the $t_n$ are not restricted to be uniform; we assume that Assumption~\eqref{ass:H} holds, as well as:
	
		{\assumption\label{ass:firm} Each operator $\mathcal{T}_i$ is firmly nonexpansive. }
		
		Then we have the following results on the iterates of Algorithm 1:
		
			\begin{theorem}\label{thh3}
Suppose that $\lambda \leq \frac{1}{8\max(1,H-1)}$. 
Then $\forall \,T\in\mathbb{N}$,
	\begin{align}
	&\frac{1}{T}\sum^{T-1}_{k=0}\Big\|\hat{x}^k - \mathcal{T}(\hat{x}^k) \Big\|^2 \leq \frac{3\|\xx - x^\star\|^2}{\lambda T}\notag\\
	&\qquad+\frac{36\lambda^2(H-1)^2}{M} \sum_{i=1}^{M}\|x^\star - \mathcal{T}_i(x^\star)\|^2.
	\end{align}
\end{theorem}


The next result gives us an explicit complexity, in terms of number of iterations sufficient to achieve $\varepsilon$-accuracy:
			
\begin{corollary}\label{cor1}
 Suppose that 
 $H\geq 2$ and that $\lambda \leq \frac{1}{8}$. Then a sufficient condition on the number $T$ of iterations  to reach $\varepsilon$-accuracy, for any $\varepsilon>0$, is
	\begin{align}
	\frac{T}{H-1} \geq \frac{24\|\hat{x}^0 - x^\star\|^2}{\varepsilon} \max\left\{2, \frac{3 \sigma}{\sqrt{\varepsilon}} \right\}.
	\end{align}\end{corollary}
	Note that as long as the target accuracy is not too high, in particular if $\varepsilon \geq \frac{9\sigma^2}{8}$, then
	$\frac{T}{H}= \mathcal{O}\left(\frac{\|\xx - x^\star\|^2}{\varepsilon}\right)$.
	If $\varepsilon < \frac{9}{8}\sigma^2 $, the communication complexity is equal to
	$\frac{T}{H}=\mathcal{O}\left(\frac{\|\xx - x^\star\|^2 \sigma}{\epsilon^{3 / 2}}\right)$.

\begin{corollary}
	Let $T\in\mathbb{N}$ and let $H\geq 1$ be such that $H\leq \frac{\sqrt{T}}{\sqrt{M}}$; set $\lambda = \frac{1}{8}\frac{\sqrt{M}}{\sqrt{T}}$. Then 
	\begin{equation}
		\frac{1}{T}\sum^{T-1}_{k=0}\Big\|\hat{x}^k - \mathcal{T}(\hat{x}^k) \Big\|^2 \!\leq \frac{24\|\xx - x^\star\|^2}{\sqrt{MT}}+\frac{3M(H-1)^2\sigma^2}{8T}.
	\end{equation}
	\end{corollary}
Hence, to get a convergence rate of $\frac{1}{\sqrt{MT}}$ we can choose the parameter $H$ as $ \mathcal{O}\left(T^{1/4}M^{-3/4}\right)$, which implies a total
	number of $\Omega\left(T^{3 / 4} M^{3 / 4}\right)$ synchronization steps. If we need a rate of $1/\sqrt{T}$, we can set a larger value $H = \mathcal{O}\left(T^{1/4}\right)$.\medskip


\begin{remark}[\textbf{Case $\boldsymbol{H=1}$}] We remark that if $H=1$, i.e.\ communication occurs after every iteration, the last term in Theorem \ref{thh3}, which depends on $H-1$, is zero. This is coherent with the fact that $x^\dagger=x^\star$ in that case, so that the algorithm converges to an exact fixed point of $\mathcal{T}$. In that sense, Theorem \ref{thh3} is tight.
\end{remark}\medskip
	\begin{remark}[\textbf{Local gradient descent (GD) case}]
	Consider that $\mathcal{T}_i(x_i^k) = x_i^k - \frac{1}{L}\nabla f_i(x_i^k) $, where each convex function $f_i$ is $L$-smooth; that is, $f_i$ is differentiable with $L$-Lipschitz continuous gradient. Then the assumptions in Theorem \ref{thh3} are satisfied and our results recover known results about Local GD for heterogeneous data as particular cases \cite{localGD}.
\end{remark}

\subsection{Linear convergence with contractive operators}


	\begin{theorem}[\textbf{Linear convergence}]\label{thh2}
			  Suppose that $t_n=nH$, for every $n\in\mathbb{N}$, and suppose that the $\mathcal{T}_i$ are all $\chi$-contractive, for some $\chi\in [0,1)$. Let $\lambda\in (0,\frac{2}{1+\chi})$ be the parameter in Algorithm 1.
		Then the the fixed point $x^\dagger$ of $\widetilde{\mathcal{T}}_\lambda$ exists and is unique, and the
		sequence $(\hat{x}^{nH})_{n\in\mathbb{N}}$ converges linearly to $x^\dagger$. More precisely, the following hold:
		
		\noindent(i) $\widetilde{\mathcal{T}}_{\lambda}$ is $\xi^H$-contractive, with
		$
		\xi=\max\big( \lambda\chi+(1-\lambda),\lambda(1+\chi)-1 \big).
		$
		
		\noindent(ii) For every $n\in\mathbb{N}$,
		\begin{equation}
		\|\hat{x}^{(n+1)H}-x^\dagger\|\leq \xi^H \|\hat{x}^{nH}-x^\dagger\|.\label{eq021}
		\end{equation}
		
		\noindent(iii) We have linear convergence with rate $\xi$: for every $n\in\mathbb{N}$,
		\begin{equation}
		\|\hat{x}^{nH}-x^\dagger\|\leq \xi^{nH} \|\hat{x}^{0}-x^\dagger\|.\label{eq022}
		\end{equation}
		
		
		
	\end{theorem}


	\noindent\textit{Proof}. For every $i=1,\ldots,M$, the operator $\lambda \mathcal{T}_i+(1-\lambda)\mathrm{Id}$ is $\xi$-contractive, with $\xi=\{ \lambda\chi+(1-\lambda)$ if $\lambda\leq 1$, $\lambda(1+\chi)-1$ else$\}$. Thus, $(\lambda \mathcal{T}_i+(1-\lambda)\mathrm{Id})^H$ is $\xi^H$ contractive. Furthermore, the average of $\xi^H$-contractive operators is $\xi^H$-contractive. The claimed properties are applications of the Banach--Picard theorem (Theorem 1.50 in \cite{bau17}).\hfill$\square$\medskip


	%
	%


\begin{remark}[\textbf{Convergence speed}]
	In the conditions of Theorem \ref{thh2}, the convergence rate $\xi$ with respect to the number of iterations is the same, whatever $H$: the distance to a fixed point is contracted by a factor of $\xi$ after every iteration, in average. The fixed point depends on $H$, however.
\end{remark} 

\begin{remark}[\textbf{Choice of $\boldsymbol{\lambda}$}]
	In the conditions of Theorem  \ref{thh2}, without further knowledge on the operators $\mathcal{T}_i$, we should set $\lambda=1$, so that $\xi=\chi$, since every other choice may slow down the convergence.
\end{remark}

Since Algorithm 1 converges linearly to $x^\dagger$, it remains to characterize the distance between $x^\dagger$ and $x^\star$. 

\begin{theorem}[\textbf{Neighborhood of the solution}]\label{thdis} In the conditions of Theorem \ref{thh2}, suppose that $\lambda=1$. So, $\xi=\chi$. Then
\begin{equation}
\|x^\dagger-x^\star\|\leq S,
\end{equation}
where
\begin{equation}
S= \frac{\xi}{1-\xi}\frac{1-\xi^{H-1}}{1-\xi^H}\frac{1}{M}\sum_{i=1}^M \|\mathcal{T}_i (x^\star)-x^\star\|.\label{eqnei}
\end{equation}
\end{theorem}
\pagebreak

\begin{remark}[\textbf{Comments on Theorem \ref{thdis}}]\ 

\textbf{(1)}\  \ If $M=1$, $\mathcal{T}_1=\mathcal{T}$, so that  $\|\mathcal{T}_1(x^\star)-x^\star\| =0$ and $S=0$, so that we recover that $x^\dagger=x^\star$, whatever $H$. In that case, the unique node and the master do not need to communicate, and the variable at the node will converge to $x^\star$. In other words, communication is irrelevant in that case.

\textbf{(2)}\ \ If $H=1$, $1-\xi^{H-1}=0$ and $S=0$, so that we recover that $x^\dagger=x^\star$.

\textbf{(3)}\ \ If $H\rightarrow +\infty$, $S$ is finite and we have
\begin{align}
S&= \frac{\xi}{1-\xi}\frac{1}{M}\sum_{i=1}^M \|\mathcal{T}_i (x^\star)-x^\star\|. \label{eqSinf}
\end{align}
This corresponds to $x^\dagger=\frac{1}{M} \sum_{i=1}^M x^\star_i$,  where $x^\star_i$ is the fixed point of $T_i$. 

\textbf{(4)}\ \ If we let $H$ vary from $1$ to $+\infty$, $S$ increases monotonically from $0$ to the value in \eqref{eqSinf}.

\textbf{(5)}\ \ In `one-shot minimization', applying $\mathcal{T}_i$ consists in going to its fixed point: $\mathcal{T}_i(x)=x_i^\star$, for every $x$. Then $\xi=0$. Hence, $S=0$, because $x^\dagger=\frac{1}{M} \sum_{i=1}^M x^\star_i = x^\star$.

\textbf{(6)}\ \ In the homogeneous case $\mathcal{T}_i=\mathcal{T}$ for every $i$, 
\begin{align*}
S&= \frac{\xi}{1-\xi}\frac{1}{M}\sum_{i=1}^M \|\mathcal{T} (x^\star)-x^\star\|=0,
\end{align*}
since $\mathcal{T} (x^\star)=x^\star$. In this case, the $M$ nodes do the same computations, so this is the same as having only one node, like in (1).

\textbf{(7)}\ \ As a direct corollary of Theorem \ref{thh2} (iii) and Theorem \ref{thdis}, we have, for every $n\in\mathbb{N}$,
\begin{align}
\|\hat{x}^{nH}-x^\star\|&\leq \xi^{nH} \|\hat{x}^{0}-x^\dagger\|+S\\
&\leq \xi^{nH} (\|\hat{x}^{0}-x^\star\|+S)+S.
\end{align}
\end{remark}



  \begin{figure*}[t]
    \centering
		\begin{tabular}{cc}
			$\!\!\!\!$\includegraphics[scale=0.2]{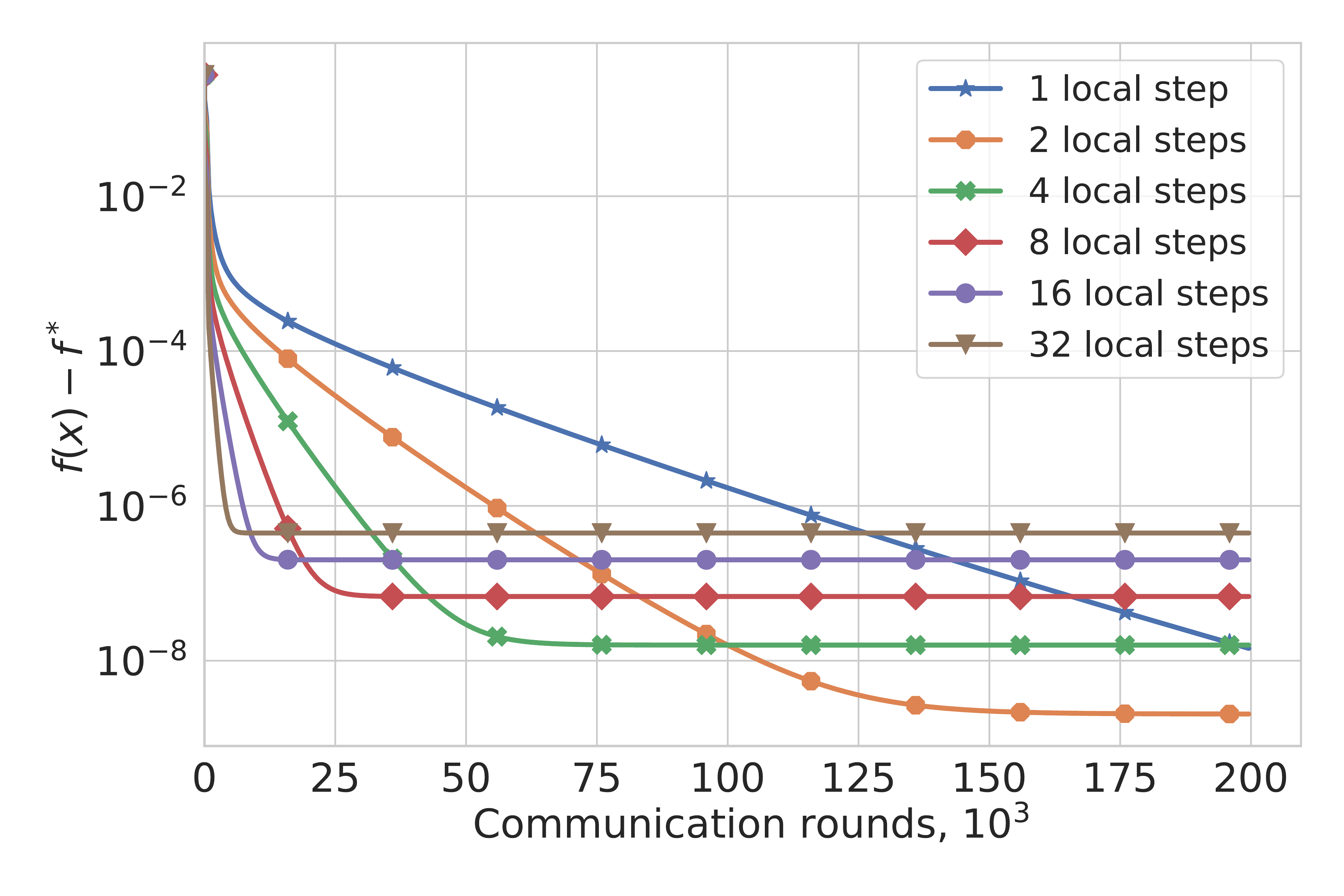}&
			$\!\!\!\!$\includegraphics[scale=0.2]{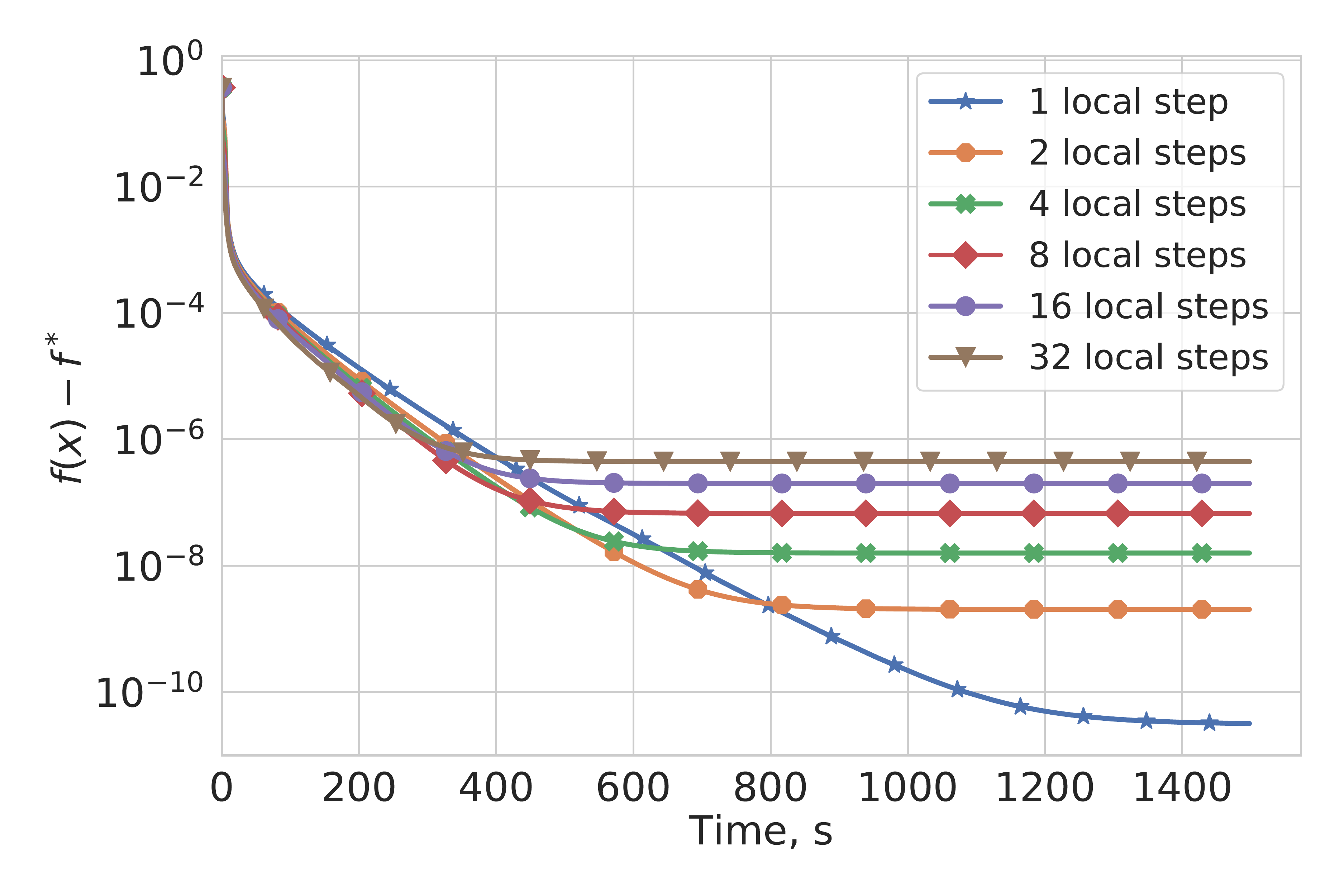}\\
			(a)&(b)
		\end{tabular}
		\begin{tabular}{c}
			$\!\!\!\!$\includegraphics[scale=0.2]{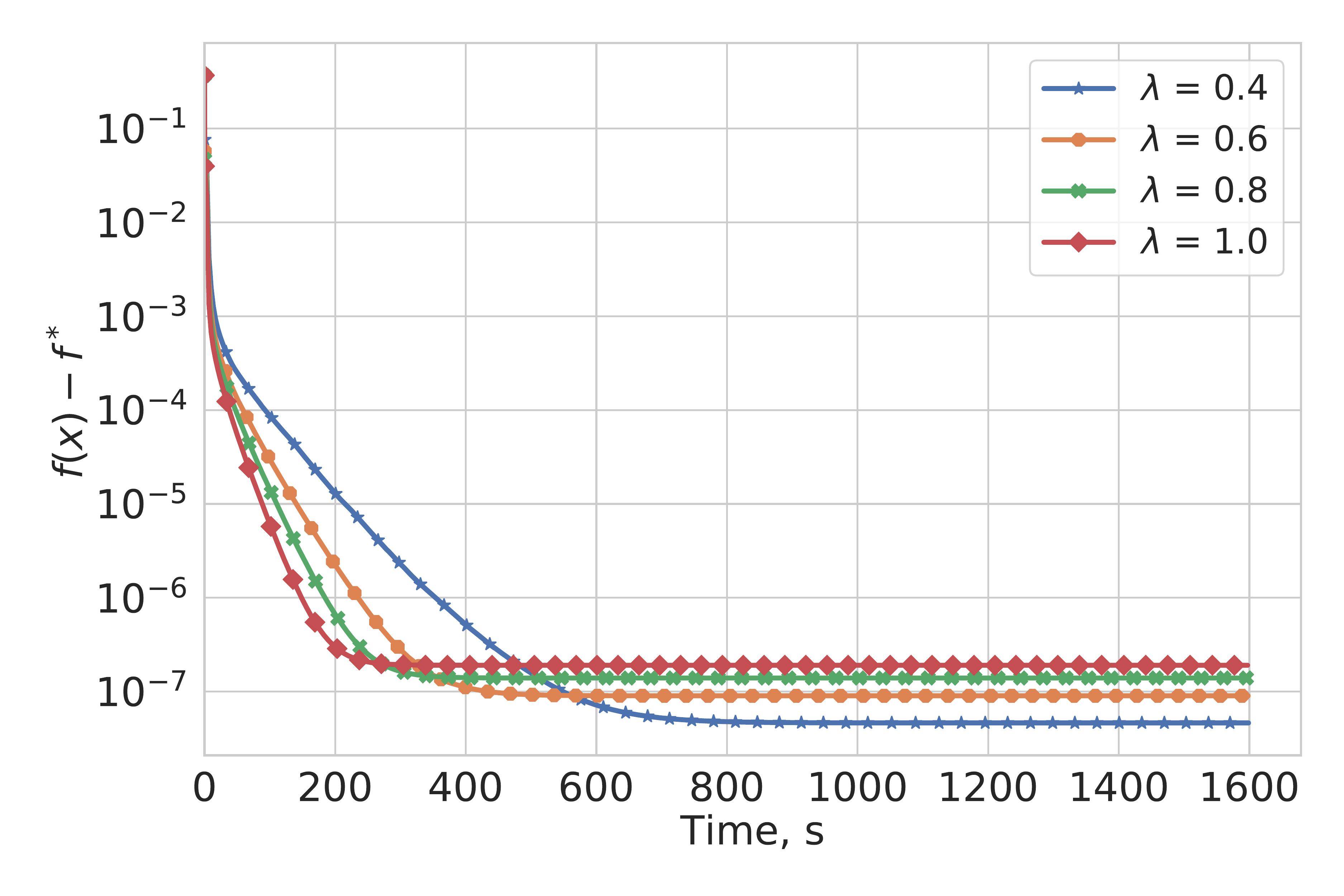}\\
			(c)
		\end{tabular}
		\caption{We analyze the convergence of Algorithm 1 with gradient descent steps, with uniform communication times $t_n=nH$;   in (a) w.r.t.\  number of communication rounds, for different values of $H$, with $\lambda=0.5$; in (b) w.r.t.\ computation time, for different values of $H$, with $\lambda=0.5$; in (c) w.r.t.\ computation time, for different values of  $\lambda$, with $H=4$.}
		\label{fig:image}
	\end{figure*}

\begin{remark}[\textbf{Local gradient descent}]\ Let us consider that each $\mathcal{T}_i:x\mapsto x - \gamma \nabla F_i(x)$, for some $L$-smooth and $\mu$-strongly convex function $F_i$, with $L\geq \mu>0$ and $0<\gamma\leq 2/(L+\mu)$. Set $\lambda=1$. Then $\xi=\chi=1-\gamma\mu$ and  $\|\mathcal{T}_i (x^\star)-x^\star\|=\gamma\|\nabla F_i(x^\star)\|$. To our knowledge, our characterization of the convergence behavior is new and improves upon state-of-the-art results \cite{localGD}, even in this case.
\end{remark}

To summarize, in presence of contractive operators, Algorithm 1 converges at the same rate as the baseline algorithm ($H=1$), up to a neighborhood of size $S$, for which we give a tight bound. 
So, if the desired accuracy $\epsilon=\|\hat{x}^k-x^\star\|$ is not lower than $S$, using local steps is the way to go, since the communication load is divided by $H$, chosen as the largest value such that $S\leq \epsilon$ in \eqref{eqnei}.
 

	\section{A Randomized Communication-Efficient Distributed Fixed-Point Method}

	Now, we propose a second loopless algorithm, where the local steps in Algorithm 1, which can be viewed as an inner loop between two communication steps, is replaced by
	a probabilistic aggregation. This yields Algorithm 2, shown above. It is communication-efficient in the following sense:  while in Algorithm 1 the number of communication rounds is divided by $H$ (or by the average of $t_n-t_{n-1}$ in the nonuniform case), in Algorithm 2 it is multiplied by the probability $p\leq 1$. Thus, $p$ plays the same role as $1/H$.

	To analyze Algorithm 2, we suppose that the operators are contractive:

	{\assumption\label{ass:s2}
		Each operator $\mathcal{T}_i$ is 
		$(1 + \rho/2)$-cocoercive \cite{bau17}, with $\rho>0$; that is, 
		there exists $\rho>0$ such that, for every $i=1,\ldots,M$ and every $x,y\in\mathbb{R}^d$,
		\begin{equation*}
		(1+\rho)\|\mathcal{T}_i(x) - \mathcal{T}_i(y)\|^2\leq\|x-y\|^2-\|x-\mathcal{T}_i(x)-y+\mathcal{T}_i(y)\|^2.
		\end{equation*}
		}
		In the particular case of gradient descent (GD) as the operator, this assumption is satisfied with $\rho>0$ for strongly convex smooth functions, see Theorem~2.1.11 in \cite{NesterovBook}.
		

	Almost sure linear convergence of Algorithm 2 up to a neighborhood is established in the next theorem:

	
  \begin{figure*}[t]
	\centering
	\begin{tabular}{cc}
		$\!\!\!\!\!\!$\includegraphics[scale=0.2]{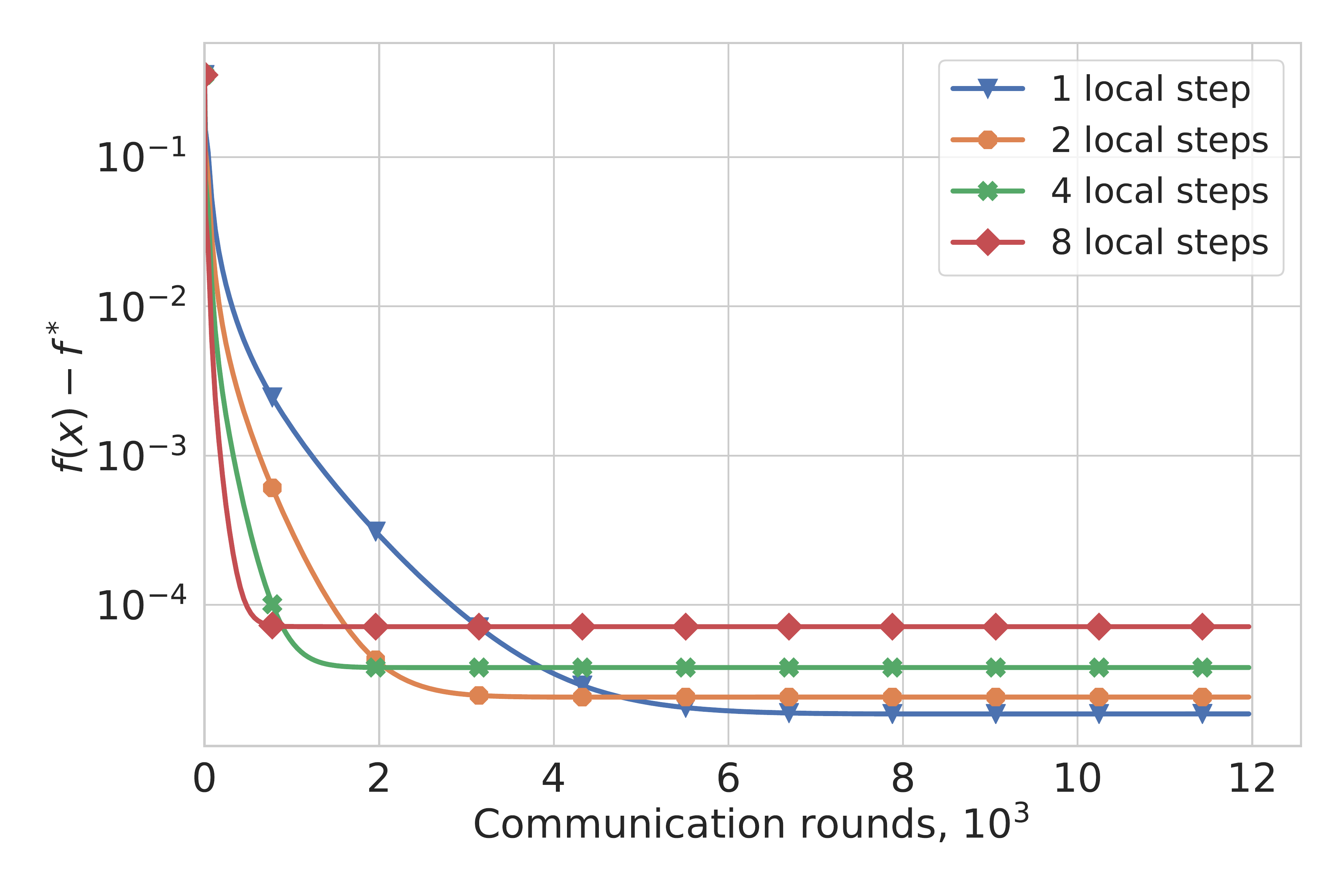}&
		$\!\!\!\!$\includegraphics[scale=0.2]{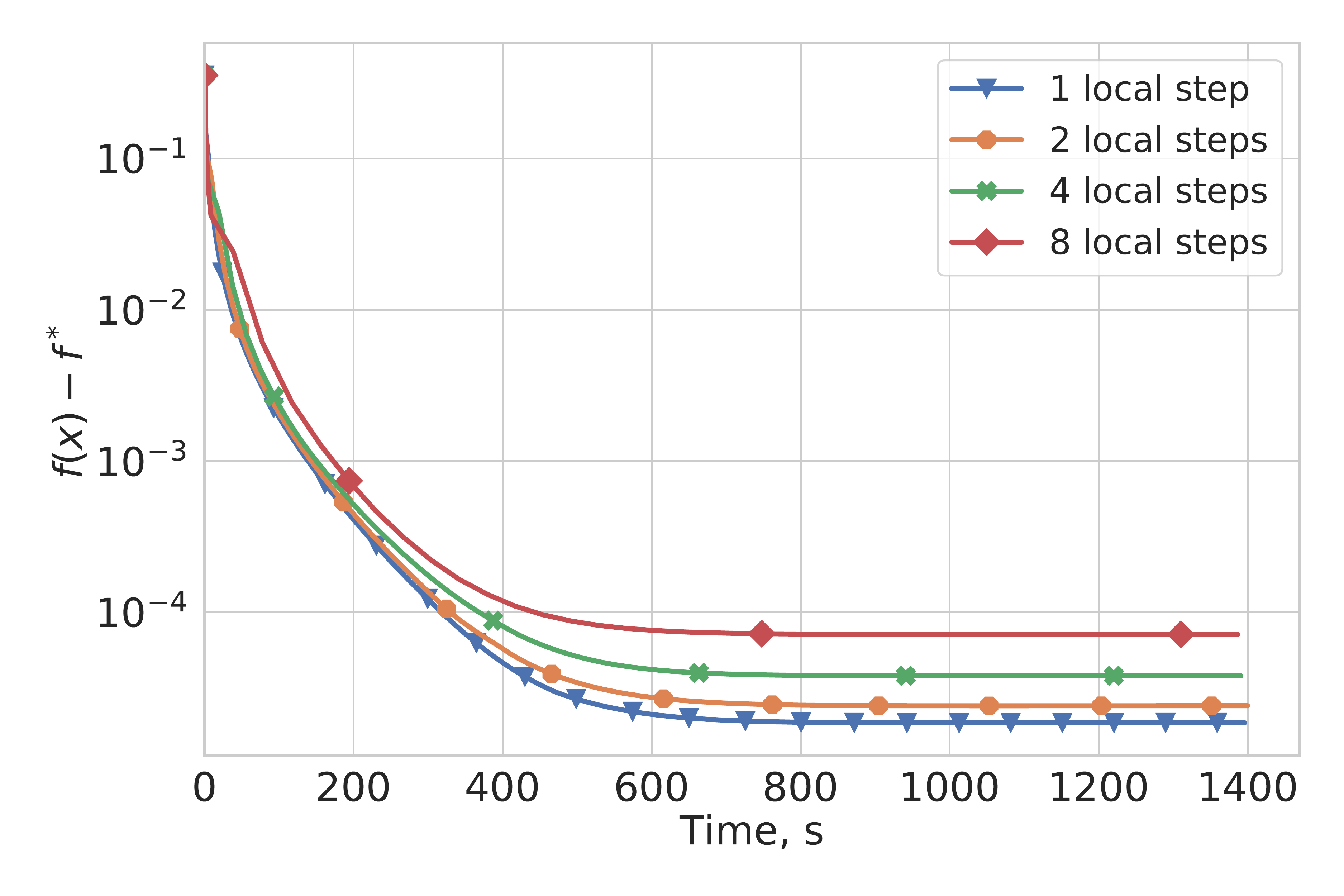}\\
		(a)&(b)
	\end{tabular}
	\begin{tabular}{c}
	$\!\!\!\!$\includegraphics[scale=0.2]{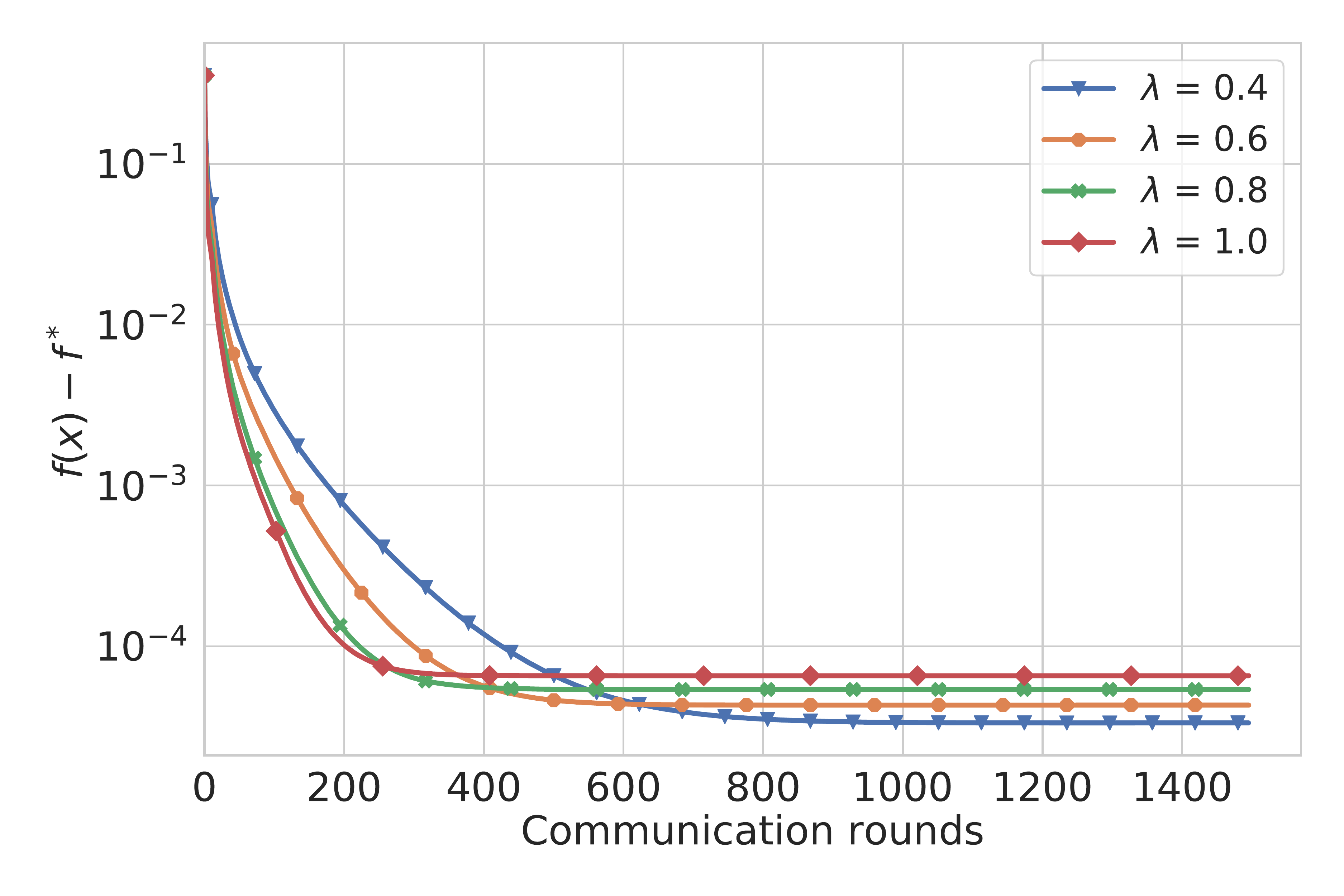}\\
	(c)
	\end{tabular}
	\caption{We analyze the convergence of Algorithm 1 with cyclic gradient descent steps, with uniform communication times $t_n=nH$;  in (a) w.r.t.\  number of communication rounds, for different values of $H$, with $\lambda=0.5$; in (b) w.r.t.\ computation time, for different values of $H$, with $\lambda=0.5$; in (c) w.r.t.\ computation time, for different values of  $\lambda$, with $H=4$.}
	\label{fig:image2}
\end{figure*}

\begin{theorem}\label{therrm2}
	Let us define the Lyapunov function: for every $k\in\mathbb{N}$,
	\begin{equation}
	\Psi^k \coloneqq     	\|\hat{x}^{k} - x^\star\|^2  + \frac{5\lambda}{p} \frac{1}{M} \sum_{i=1}^{M}\left\|x_{i}^{k}-\hat{x}^{k}\right\|^{2} 
	\end{equation}
	Then, under Assumption \ref{ass:s2} and if $\lambda<\frac{p}{15}$,  we have, for every $k\in\mathbb{N}$,
	\begin{align}
	\mathbb{E}\Psi^{k}&\leq \left(1-\min\left(\frac{\lambda\rho}{1+\rho},\frac{p}{5}\right)\right)^k\Psi^0\notag\\
	&\quad+\frac{150}{\min\left(\frac{\lambda\rho}{1+\rho},\frac{p}{5}\right)p^2}\lambda^3\sigma^2,
	\end{align}
	where $\sigma^2 \coloneqq \frac{1}{M} \sum_{i=1}^{M}\|x^\star - \mathcal{T}_i(x^\star) \|^2$ and $\mathbb{E}$ denotes the expectation.
\end{theorem}

Since the previous theorem may be difficult to analyze, the next results gives a bound to reach $\varepsilon$-accuracy in  in Algorithm \ref{alg2}:
\begin{corollary}\label{cor2}
Under Assumption \ref{ass:s2} and if $\lambda<\frac{p}{15}$, for any $\varepsilon>0$, $\varepsilon$-accuracy  is reached after $T$ iterations, with
\begin{align}
T &\geq \max \left\{\frac{15(1 + \rho)}{\rho p}, \frac{18 \sigma (1 + \rho)^\frac13}{p \rho^\frac32 \varepsilon^\frac12}, \frac{40 \sigma^\frac23(1 + \rho)}{p \rho \varepsilon^\frac13} \right\}\notag\\
&\quad{}\times\log \frac{2 \Psi_0}{\varepsilon}.
\end{align}
\end{corollary}


	\begin{figure*}[t]
	\centering
		\begin{tabular}{cc}
			$\!\!\!\!$\includegraphics[scale=0.2]{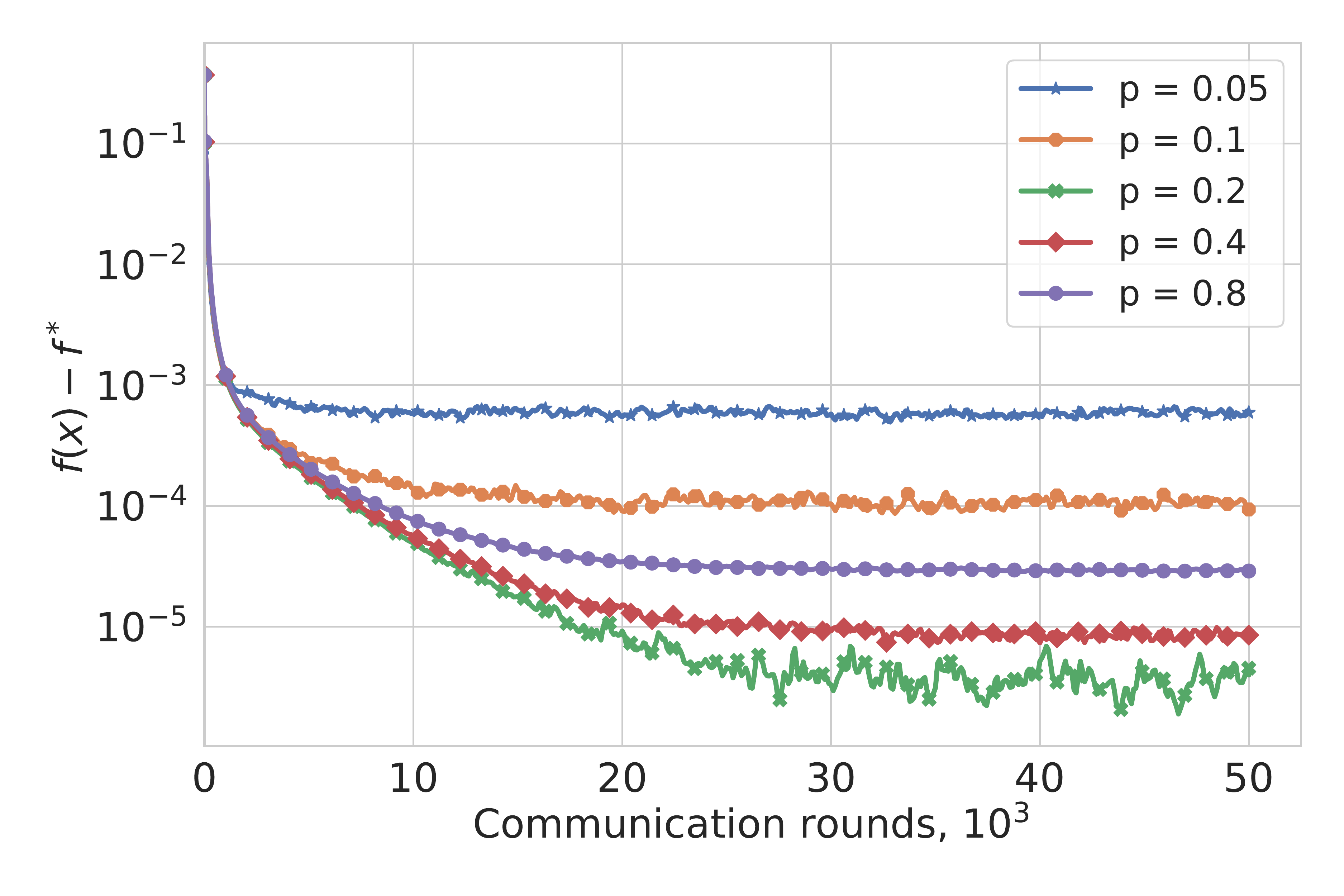}&
			$\!\!\!\!$\includegraphics[scale=0.2]{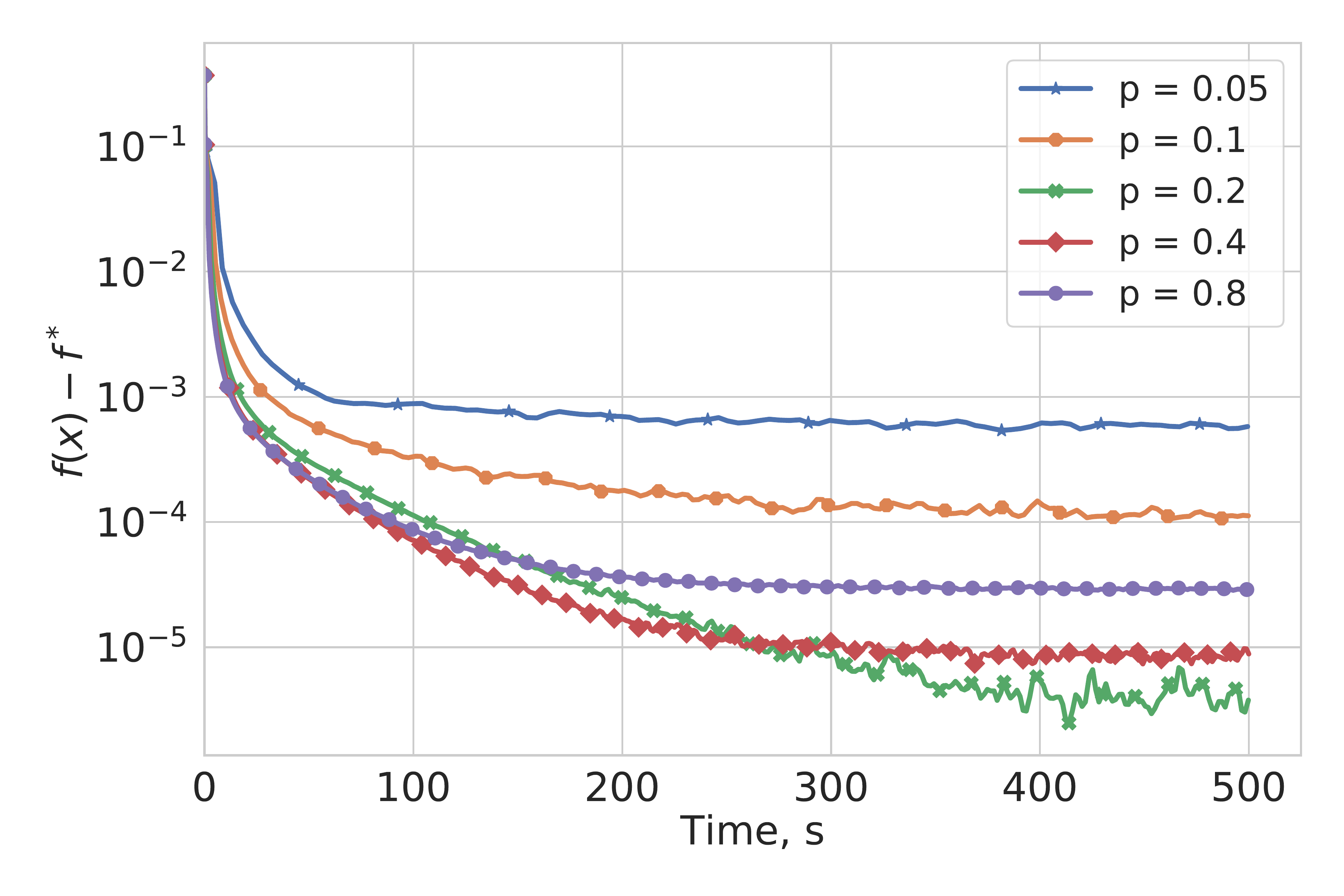}\\
			(a)&(b)
		\end{tabular}
		\begin{tabular}{c}
		$\!\!\!\!$\includegraphics[scale=0.2]{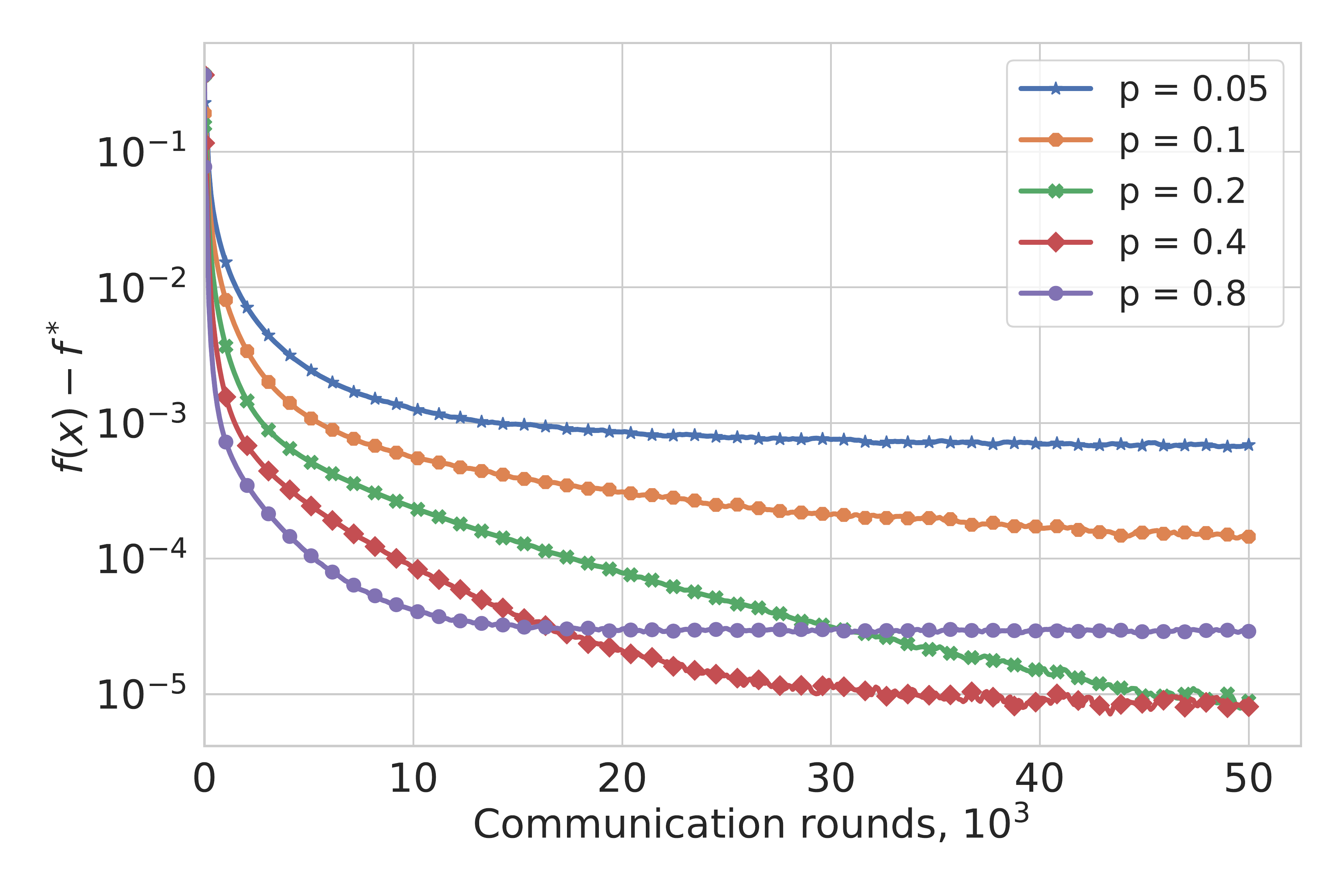}\\
		(c)
		\end{tabular}
		\caption{We analyze the convergence of Algorithm 2 with gradient descent steps,  with $\lambda=0.5$; in (a)
			with the same gradient stepsizes, w.r.t.\ number of communication rounds, for different values of $p$; in (b)
			with the same gradient stepsizes, w.r.t.\ computation time, for different values of $p$; in (c)
			with gradient stepsizes proportional to $p$, w.r.t.\ number of communication rounds, for different values of $p$. }
		\label{fig:image3}
	\end{figure*}
	

\begin{figure*}[t]
\centering
	\begin{tabular}{cc}
		$\!\!\!\!$\includegraphics[scale=0.2]{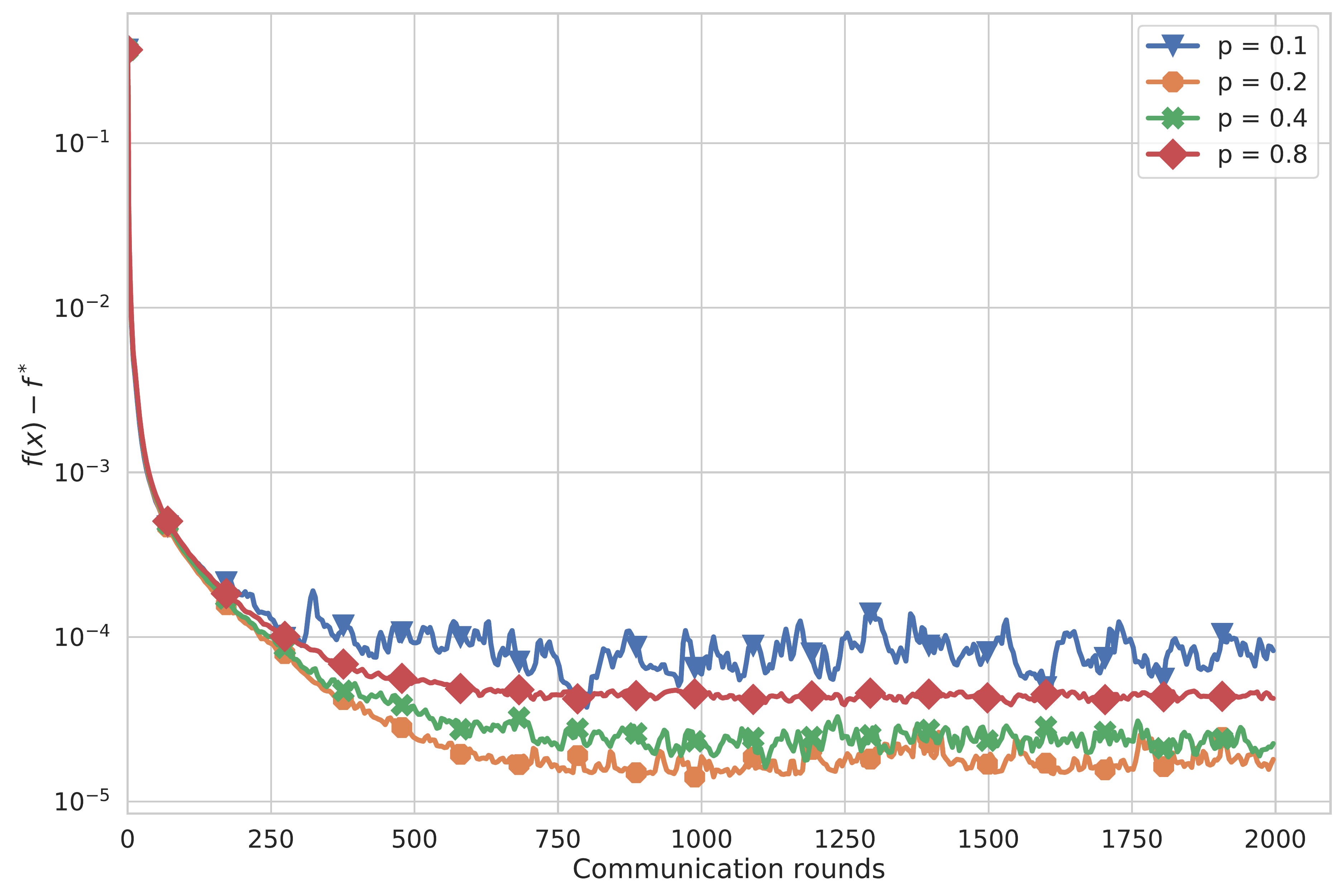}&
		$\!\!\!\!$\includegraphics[scale=0.2]{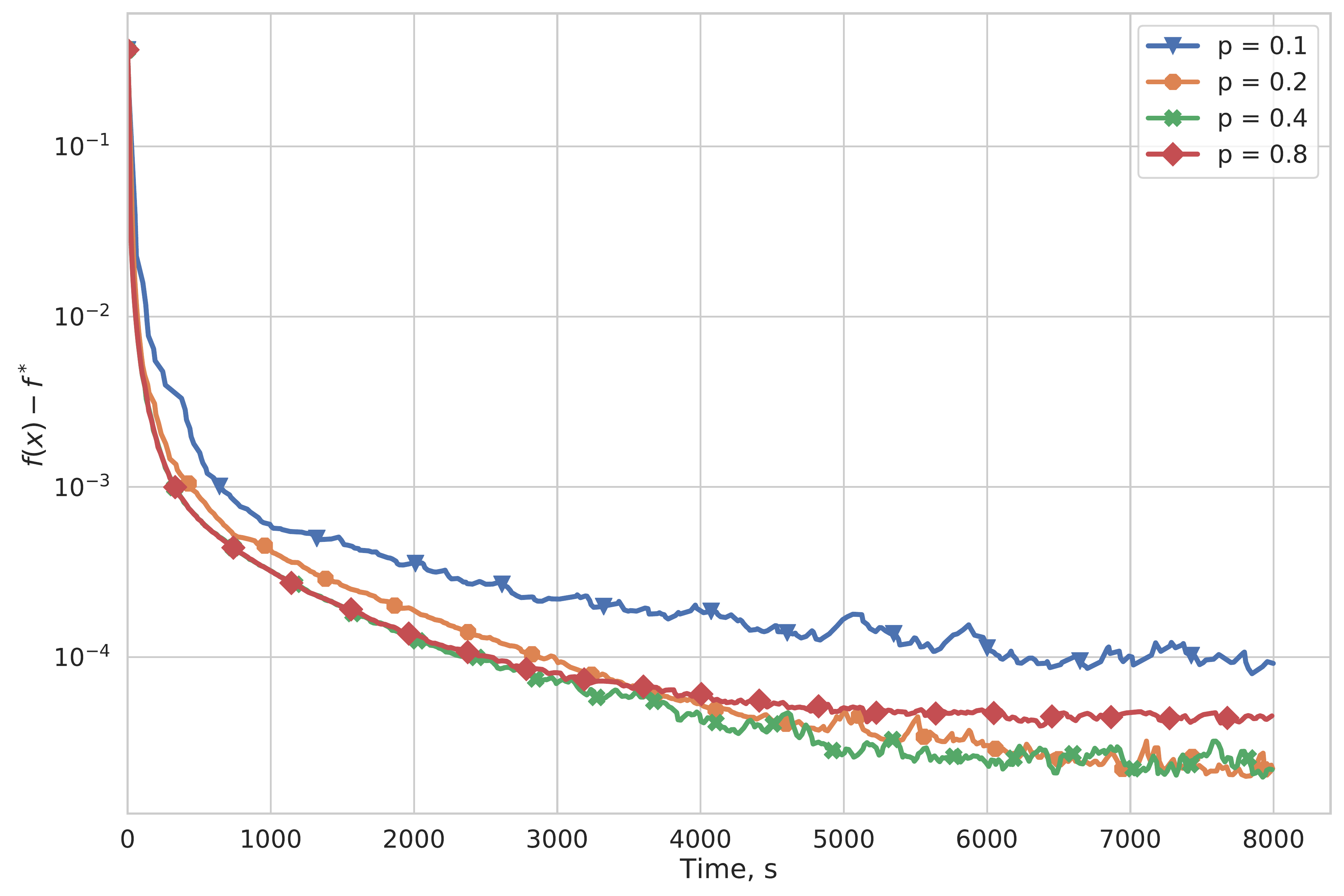}\\
	(a)&(b)
	\end{tabular}	
	\begin{tabular}{c}
		$\!\!\!\!$\includegraphics[scale=0.2]{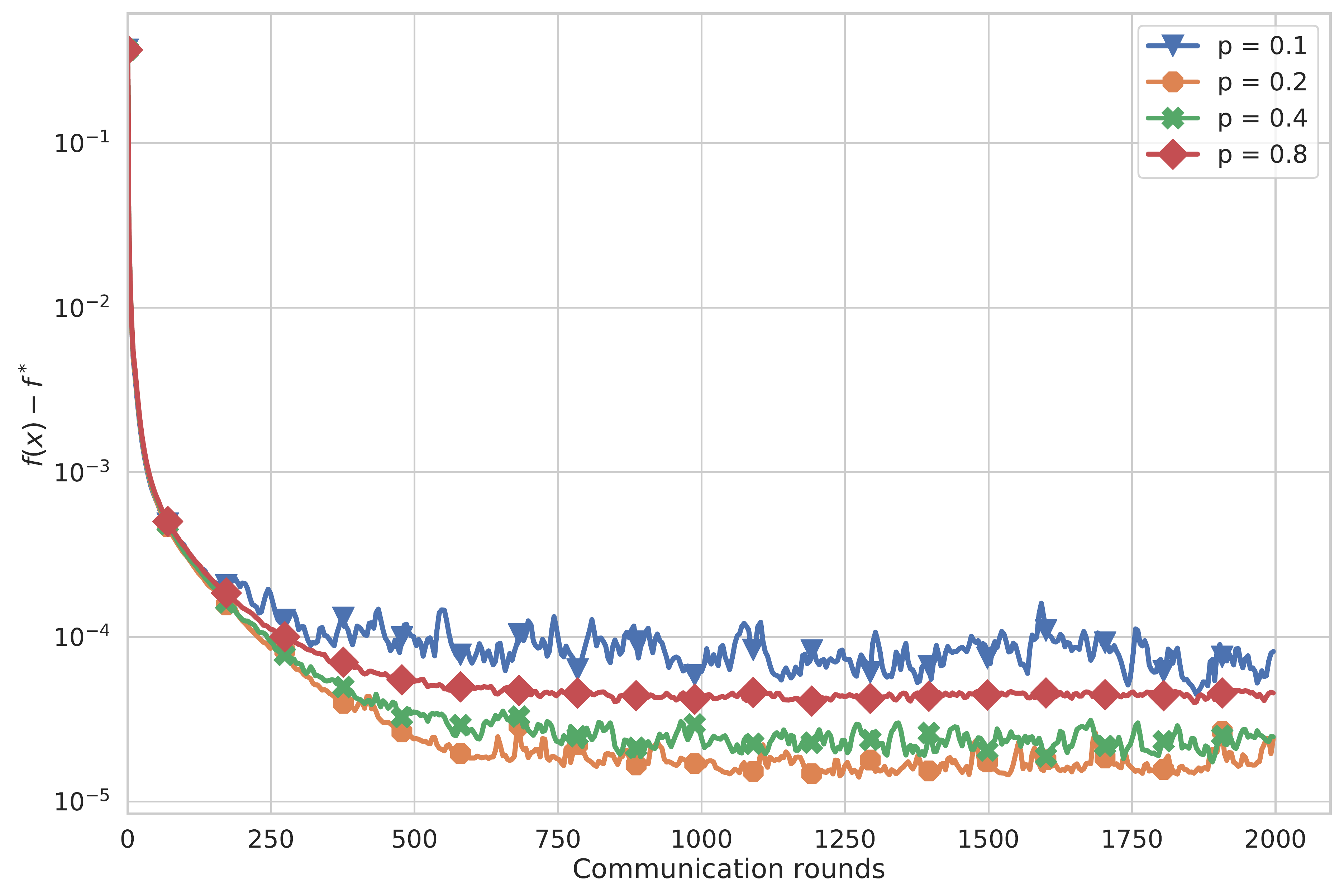}\\
		(c)
	\end{tabular}
	\caption{We analyze the convergence of Algorithm 2 with cyclic gradient descent steps,  with $\lambda=0.5$; in (a)
		with the same gradient stepsizes, w.r.t.\ number of communication rounds, for different values of the $p$; in (b) same as in (a), but w.r.t.\ computation time; in (c)
		with gradient stepsizes proportional to $p$, w.r.t.\ number of communication rounds, for different values of $p$. 
		}
	\label{fig:image4}
\end{figure*}

	\section{Experiments}
	\textbf{Model}\ \ 
	Although our approach can be applied more broadly, we focus on logistic regression, since this is one of the most important models for classification. The corresponding objective function takes the following form:
	\begin{equation*}
	f(x)=\frac{1}{n} \sum_{i=1}^{n} \log \left(1+\exp \left(-b_{i} a_{i}^{\top} x\right)\right)+\frac{\kappa}{2}\|x\|^{2},
	\end{equation*} where $a_i \in \mathbb{R}^d$ and $b_i \in \left\lbrace -1, +1\right\rbrace $ are the data samples.

	\textbf{Datasets}\ \ 
	We use the 'a9a' and 'a4a' datasets
	from the
	LIBSVM library  and we set $\kappa$
	to be $\frac{L}{n}$, where $n$ is the size of the dataset and $L$ is a Lipschitz constant of
	the first part of $\nabla f$, without regularization.

	\textbf{Hardware and software}\ \ We implemented all algorithms in Python using the package MPI4PY, in order to run the code on a truly parallel architecture. All methods were evaluated on a computer with an Intel(R) Xeon(R) Gold 6146 CPU at 3.20GHz, having 24 cores. The cores are connected to 2 sockets, with 12 cores for each of them.

	\subsection{Local gradient descent}
	We consider gradient descent (GD) steps as the operators.
	That is, we consider the problem of minimizing the finite sum:
	\begin{equation}
	f(x) = \frac{1}{M}\sum^M_{i=1}f_i(x),\label{eqf1}
	\end{equation}
	where each function $f_i$ is convex and $L$-smooth. We set
	$\mathcal{T}_i(x_i^k) \coloneqq x^k_i - \frac{1}{L}\nabla f_i(x_i^k)$.
	We use $\frac{1}{L}$ as the stepsize, so that each $\mathcal{T}_i$ is firmly nonexpansive. 
	The results of Algorithms 1 and 2 are illustrated in Figures 1 and 3, respectively.

	\subsection{Local cycling GD}
	In this section, we consider another operator, which is cycling GD.
	So, we consider minimizing the same function as in \eqref{eqf1}, but this time each
	function $f_i$ is also a finite sum:
	$f_i = \frac{1}{N}\sum_{j=1}^{N}f_{ij}$.
	Instead of applying full gradient steps, we apply $N$ element-wise gradient steps, in the sequential order of the data points. Thus,
	\begin{equation*}
	\mathcal{T}_i(x_i^k) \coloneqq S_{i1}(S_{i2}(\ldots S_{in}(x_i^k))),
	\end{equation*}
	where
	$S_{ij} : y \mapsto y - \frac{1}{NL}\nabla f_{ij}$.
	If, for each $i$, all functions $f_{ij}$ have the same minimizer $x^\star_i$, then this joint minimizer is a fixed point of $\mathcal{T}_i$. Also, these operators can be shown to be firmly nonexpansive. The results of Algorithms 1 and 2 are illustrated in Figures 2 and 4, respectively.


	\subsection{Results}
	We observe a very tight match between our theory and the numerical results. 
	As can be seen, the larger the value of the parameters $H$ and $\lambda$, the faster the convergence at the beginning, but the larger the radius of the neighborhood. In terms of computational time, there is no big advantage, since the experiments were run on a single machine and the communication time was negligible. But in a distributed setting where communication is slow, our approach has a clear advantage. We can also observe the absence of oscillations. 
	Hence, there is a clear advantage of
	local methods when only limited accuracy is required.
	
In the experiment with cyclic GD, the algorithm converges only to a neighbourhood of the ideal solution, even when 1 local step is used. This happens because the assumption of a joint minimizer for all $i$ is not satisfied here. 	However, since the operators are firmly nonexpansive, we have convergence to a fixed point. The convergence of Algorithm~1 is illustrated with respect to the relaxation parameter $\lambda$. If $\lambda$ is small, convergence is slower, but the algorithm converges to a point closer to the true solution $x^\star$. 
	In Figure~4, we further illustrate the behavior of Algorithm~2 with respect to the probability $p$, for cyclic gradient descent. We can see that the fastest and most accurate convergence is obtained for an intermediate value of $p$, here $p=0.2$.

	The experiments with Algorithm 2 show that, with a low probability $p$ of update, the neighborhood is substantially larger; however, with $p$ increasing, the convergence in terms of communication rounds becomes worse. Therefore, with careful selection of the probability parameter,  a significant advantage can be obtained.



\pagebreak

	\section{Conclusion}
	
	We have proposed two strategies to reduce the communication burden in a generic distributed setting, where a fixed point of an average of operators is sought. We have shown that they improve the convergence speed, while achieving the goal of reducing the communication load. 
	At convergence, only an approximation of the ideal fixed point is attained, but if medium accuracy is sufficient, 
	the proposed algorithms are particularly adequate.
	
	In future work, we will generalize the setting to randomized fixed-point operators, to  generalize  stochastic gradient descent approaches. We will also investigate compression \cite{GDCI,Chraibi2019DistributedFP} of the communicated variables, with or without variance reduction, in combination with locality.

\section*{Acknowledgements}

Part of this work was done while the first author was an intern at KAUST.

	
	

\bibliographystyle{icml2020}
	\bibliography{biblio2}

\begin{thebibliography}{19}
\providecommand{\natexlab}[1]{#1}
\providecommand{\url}[1]{\texttt{#1}}
\expandafter\ifx\csname urlstyle\endcsname\relax
  \providecommand{\doi}[1]{doi: #1}\else
  \providecommand{\doi}{doi: \begingroup \urlstyle{rm}\Url}\fi

\bibitem[Bauschke \& Combettes(2017)Bauschke and Combettes]{bau17}
Bauschke, H.~H. and Combettes, P.~L.
\newblock \emph{Convex Analysis and Monotone Operator Theory in Hilbert
  Spaces}.
\newblock Springer, New York, 2nd edition, 2017.

\bibitem[Bauschke et~al.(2011)Bauschke, Burachik, Combettes, Elser, Luke, and
  Wolkowicz]{bau11a}
Bauschke, H.~H., Burachik, R.~S., Combettes, P.~L., Elser, V., Luke, D.~R., and
  Wolkowicz, H. (eds.).
\newblock \emph{Fixed-Point Algorithms for Inverse Problems in Science and
  Engineering}.
\newblock Springer, 2011.

\bibitem[Chraibi et~al.(2019)Chraibi, Khaled, Kovalev, Richt{\'a}rik, Salim,
  and Tak\'{a}\v{c}]{Chraibi2019DistributedFP}
Chraibi, S., Khaled, A., Kovalev, D., Richt{\'a}rik, P., Salim, A., and
  Tak\'{a}\v{c}, M.
\newblock Distributed fixed point methods with compressed iterates.
\newblock \emph{preprint ArXiv:1912.09925}, 2019.

\bibitem[Combettes \& Woodstock(2020)Combettes and Woodstock]{com20}
Combettes, P.~L. and Woodstock, Z.~C.
\newblock A fixed point framework for recovering signals from nonlinear
  transformations.
\newblock preprint arXiv:2003.01260, 2020.

\bibitem[Combettes \& Yamada(2015)Combettes and Yamada]{comy15}
Combettes, P.~L. and Yamada, I.
\newblock Compositions and convex combinations of averaged nonexpansive
  operators.
\newblock \emph{Journal of Mathematical Analysis and Applications},
  425\penalty0 (1):\penalty0 55--70, 2015.

\bibitem[Davis \& Yin(2016)Davis and Yin]{dav16}
Davis, D. and Yin, W.
\newblock Convergence rate analysis of several splitting schemes.
\newblock In Glowinski, R., Osher, S.~J., and Yin, W. (eds.), \emph{Splitting
  Methods in Communication, Imaging, Science, and Engineering}, pp.\  115--163,
  Cham, 2016. Springer International Publishing.

\bibitem[Haddadpour \& Mahdavi(2019)Haddadpour and
  Mahdavi]{haddadpour2019convergence}
Haddadpour, F. and Mahdavi, M.
\newblock On the convergence of local descent methods in federated learning.
\newblock \emph{preprint arXiv:1910.14425}, 2019.

\bibitem[Khaled \& Richt\'{a}rik(2019)Khaled and Richt\'{a}rik]{GDCI}
Khaled, A. and Richt\'{a}rik, P.
\newblock Gradient descent with compressed iterates.
\newblock In \emph{NeurIPS Workshop on Federated Learning for Data Privacy and
  Confidentiality}, 2019.

\bibitem[Khaled et~al.(2019)Khaled, Mishchenko, and Richt\'{a}rik]{localGD}
Khaled, A., Mishchenko, K., and Richt\'{a}rik, P.
\newblock First analysis of local {GD} on heterogeneous data.
\newblock In \emph{NeurIPS Workshop on Federated Learning for Data Privacy and
  Confidentiality}, 2019.

\bibitem[Khaled et~al.(2020)Khaled, Mishchenko, and
  Richt\'{a}rik]{localSGD-AISTATS2020}
Khaled, A., Mishchenko, K., and Richt\'{a}rik, P.
\newblock Tighter theory for local {SGD} on identical and heterogeneous data.
\newblock In \emph{The 23rd International Conference on Artificial Intelligence
  and Statistics (AISTATS 2020)}, 2020.

\bibitem[Kone\v{c}n\'{y} et~al.(2016)Kone\v{c}n\'{y}, McMahan, Yu,
  Richt\'{a}rik, Suresh, and Bacon]{ja2016}
Kone\v{c}n\'{y}, J., McMahan, H.~B., Yu, F.~X., Richt\'{a}rik, P., Suresh,
  A.~T., and Bacon, D.
\newblock Federated learning: Strategies for improving communication
  efficiency.
\newblock In \emph{NIPS Workshop on Private Multi-Party Machine Learning},
  2016.

\bibitem[Lessard et~al.(2016)Lessard, Recht, and Packards]{les16}
Lessard, L., Recht, B., and Packards, A.
\newblock Analysis and design of optimization algorithms via integral quadratic
  constraints.
\newblock \emph{SIAM J. Optim.}, 26\penalty0 (1):\penalty0 57--95, 2016.

\bibitem[Ma et~al.(2017)Ma, Kone{\v{c}}n{\'y}, Jaggi, Smith, Jordan,
  Richt{\'a}rik, and Tak{\'a}{\v{c}}]{ma2017distributed}
Ma, C., Kone{\v{c}}n{\'y}, J., Jaggi, M., Smith, V., Jordan, M.~I.,
  Richt{\'a}rik, P., and Tak{\'a}{\v{c}}, M.
\newblock Distributed optimization with arbitrary local solvers.
\newblock \emph{Optimization Methods and Software}, 32\penalty0 (4):\penalty0
  813--848, 2017.

\bibitem[McMahan et~al.(2017)McMahan, Moore, Ramage, Hampson, and {Ag\"{u}era y
  Arcas}]{FL2017-AISTATS}
McMahan, H.~B., Moore, E., Ramage, D., Hampson, S., and {Ag\"{u}era y Arcas},
  B.
\newblock Communication-efficient learning of deep networks from decentralized
  data.
\newblock In \emph{Proceedings of the 20th International Conference on
  Artificial Intelligence and Statistics (AISTATS)}, 2017.

\bibitem[Nesterov(2004)]{NesterovBook}
Nesterov, Y.
\newblock \emph{Introductory lectures on convex optimization: a basic course}.
\newblock Kluwer Academic Publishers, 2004.

\bibitem[Pesquet \& Repetti(2015)Pesquet and Repetti]{pes15}
Pesquet, J.-C. and Repetti, A.
\newblock A class of randomized primal-dual algorithms for distributed
  optimization.
\newblock \emph{J. Nonlinear Convex Anal.}, 12\penalty0 (16), December 2015.

\bibitem[Richt{\'a}rik \& Tak{\'a}{\v{c}}(2014)Richt{\'a}rik and
  Tak{\'a}{\v{c}}]{ric14}
Richt{\'a}rik, P. and Tak{\'a}{\v{c}}, M.
\newblock Iteration complexity of randomized block-coordinate descent methods
  for minimizing a composite function.
\newblock \emph{Math. Program.}, 144\penalty0 (1--2):\penalty0 1--38, April
  2014.

\bibitem[Stich(2019)]{Stich2018}
Stich, S.~U.
\newblock {Local SGD Converges Fast and Communicates Little}.
\newblock In \emph{International Conference on Learning Representations}, 2019.

\bibitem[Yu(2013)]{yu13}
Yu, Y.-L.
\newblock On decomposing the proximal map.
\newblock In \emph{Proc. of 26th Int. Conf. Neural Information Processing
  Systems (NIPS)}, pp.\  91--99, 2013.

\end{thebibliography}

	\clearpage

\onecolumn
\appendix

	\part*{Supplementary material}

	

	
	\section{Notations and Basic Facts}\label{seca1}
	\subsection{Notations}\label{not}
	Let $\mathcal{T}_1,\mathcal{T}_2,\dots,\mathcal{T}_n$ be operators on $\mathbb{R}^d$.
	
	Let us list here the notations used in the paper and the Appendix:
	\begin{align*}
	\mathcal{T}(x)=\frac{1}{M} \sum_{i=1}^{M} \mathcal{T}_{i}(x) &- \textbf{ averaging operator},\\
	  x^\star = \mathcal{T}(x^\star)  &- \textbf{ fixed point},\\
	     \hat{x}^{k} = \frac{1}{M} \sum_{i=1}^{M} x_{i}^{k} &- \textbf{ mean point},\\
	\sigma^2 = \frac{1}{M} \sum_{i=1}^{M}\|g_i(x^\star)\|^2 &- \textbf{ variance for locality},\\
	V_{k} = \frac{1}{M} \sum_{i=1}^{M}\left\|x_{i}^{k}-\hat{x}^{k}\right\|^{2} &- \textbf{ deviation from average},\notag\\
	g_{i}(x) = x - \mathcal{T}_i(x) &- \textbf{ local residual},\\
	\hat{g}^{k} = \hat{x}^k - \frac{1}{M}\sum^{M}_{i=1}\mathcal{T}_i(x^k_{i}) &- \textbf{residual for mean point},\notag\\
	\rho &- \textbf{ contraction parameter},\\
	\lambda &- \textbf{relaxation parameter},\notag\\[3mm]
	H &- \textbf{ bound for the number of local steps in Alg.~\ref{alg}},\\
	p &- \textbf{probability of communication in Alg.~\ref{alg2}}.
	\end{align*}  
	
	The value $V_k$ measures the deviation of
	the iterates from their average. This value is crucial for the convergence analysis. The values $g_i(x^k)$ and $\hat{g}^k$ can be viewed as analogues of the gradient and the average gradient in our more general setting. The value $\sigma^2$ serves as a measure of
	variance adapted to methods with local steps.
	
	\subsection{Basic Facts}
	\textbf{Jensen’s inequality}. For any convex function $f$ and any vectors $x^1,\ldots x^M$ we have
	\begin{equation}
	f\left(\frac{1}{M} \sum_{m=1}^{M} x^{m}\right) \leq \frac{1}{M} \sum_{m=1}^{M} f\left(x^{m}\right).
	\end{equation}   
	In particular, with $f(x) = \|x\|^2$, we obtain
	\begin{equation}\label{jensen}
	\left\|\frac{1}{M} \sum_{m=1}^{M} x_{m}\right\|^{2} \leq \frac{1}{M} \sum_{m=1}^{M}\left\|x_{m}\right\|^{2}.
	\end{equation}
	\textbf{Facts from linear algebra}. \label{young_1_2}\\
	We will use the following important properties:
	\begin{equation}
	\label{young1}
	\|x+y\|^{2} \leq 2\|x\|^{2}+2\|y\|^{2}, \text { for every } x, y \in \mathbb{R}^{d},
	\end{equation}
	\begin{equation}
	\label{young3}
	\|x+y\|^{2} \geq \frac{1}{2}\|y\|^{2}-\|x\|^{2}, \text { for every } x, y \in \mathbb{R}^{d},
	\end{equation}
	\begin{equation}
	\label{young2}
	2\langle a, b\rangle \leq \zeta\|a\|^{2}+\zeta^{-1}\|b\|^{2} \text { for all } a, b \in \mathbb{R}^{d} \text { and } \zeta>0,
	\end{equation}
	\begin{equation}
	\label{moment}
	\frac{1}{M} \sum_{m=1}^{M}\left\|X_{m}\right\|^{2}=\frac{1}{M} \sum_{m=1}^{M}\left\|X_{m}-\frac{1}{M} \sum_{i=1}^{M} X_{i}\right\|^{2}+\left\|\frac{1}{M} \sum_{m=1}^{M} X_{m}\right\|^{2}.
	\end{equation}
	
	\textbf{Firm nonexpansiveness} An operator $\mathcal{T}$ is said to be firmly nonexpansive if it is $1/2$-averaged. 
	Equivalently, for every $x$ and $y\in\mathbb{R}^d$,
		\begin{equation}
		\|\mathcal{T}(x) - \mathcal{T}(y)\|^2\leq \|x-y\|^2 - \|\mathcal{T}(x) - x - \mathcal{T}(y) + y\|^2.
		\end{equation}

	\subsection{Technical lemmas}
	\textbf{Technical Lemma 1}. 
	If $\mathcal{T}$ is firmly nonexpansive, then
	\begin{equation}
	\langle \mathcal{T}(x) - x - \mathcal{T}(y) + y, x - y \rangle \leq
	- \|x - \mathcal{T}(x) - y + \mathcal{T}(y) \|^2
	\end{equation}
	and
	\begin{equation}
	\|\mathcal{T}(x) - \mathcal{T}(y)\|^2 \leq \|x-y\|^2.
	\end{equation}
	\begin{proof}
		\begin{align*}
		\|\mathcal{T}(x) - \mathcal{T}(y)\|^2 &= \|x-y\|^2+2\langle   \mathcal{T}(x) - x - \mathcal{T}(y) + y, x-y \rangle + \| \mathcal{T}(x) - x - \mathcal{T}(y) + y \|^2\\&
		\leq \|x-y\|^2 - \|\mathcal{T}(x) - x - \mathcal{T}(y) + y\|^2\\&  = \|x-y\|^2 - \|x - \mathcal{T}(x) - y + \mathcal{T}(y) \|^2.
		\end{align*}
		We have
		$$\|x-y\|^2+2\langle   \mathcal{T}(x) - x - \mathcal{T}(y) + y, x-y \rangle + \| \mathcal{T}(x) - x - \mathcal{T}(y) + y \|^2 \leq \|x-y\|^2 - \|x - \mathcal{T}(x) - y + \mathcal{T}(y) \|^2. $$
		So,
		$$2\langle   \mathcal{T}(x) - x - \mathcal{T}(y) + y, x-y \rangle + \| \mathcal{T}(x) - x - \mathcal{T}(y) + y \|^2 \leq - \|x - \mathcal{T}(x) - y + \mathcal{T}(y) \|^2, $$
		$$2\langle   \mathcal{T}(x) - x - \mathcal{T}(y) + y, x-y \rangle  \leq -2 \|x - \mathcal{T}(x) - y + \mathcal{T}(y) \|^2, $$
		$$\langle   \mathcal{T}(x) - x - \mathcal{T}(y) + y, x-y \rangle  \leq - \|x - \mathcal{T}(x) - y + \mathcal{T}(y) \|^2. $$%
	\end{proof}
	\textbf{Technical Lemma 2}. 
	Let $\rho>0$. Let $\mathcal{T}$ be a contractive and firmly nonexpansive operator; that is, for every $x,y\in\mathbb{R}^d$,
	\begin{equation}
	(1+\rho)\|\mathcal{T}(x) - \mathcal{T}(y)\|^2\leq\|x-y\|^2-\|x-\mathcal{T}(x)-y+\mathcal{T}(y)\|^2.
	\end{equation}
	Then
	\begin{equation*}
	\langle x-y,\mathcal{T}_i(x)-x +y - \mathcal{T}_i(y) \rangle \leq - \left(\frac{\rho}{2(1+\rho)}\|x-y\|^2+\frac{2+\rho}{2(1+\rho)}\|x-\mathcal{T}_i(x) - y + \mathcal{T}_i(y)\|^2\right).
	\end{equation*}
	\begin{proof}
		\begin{align*}
		\|\mathcal{T}_i(x) - \mathcal{T}_i(y)\|^2 &= \|x-y - x + \mathcal{T}_i(x) + y - \mathcal{T}_i(y)\|^2\\&
		= \|x-y\|^2-2\langle x-y,x-\mathcal{T}_i(x) - y + \mathcal{T}_i(y)\rangle+\|x-\mathcal{T}_i(x) - y + \mathcal{T}_i(y) \|^2.
		\end{align*}
		We have
		\begin{align*}
		(1+\rho)\|\mathcal{T}_i(x) - \mathcal{T}_i(y)\|^2 &=(1+\rho)\|x-y\|^2+(1+\rho)\|x-\mathcal{T}_i(x) - y + \mathcal{T}_i(y)\|^2 \\
		&\quad{}- 2(1+\rho)\langle x-y,x-\mathcal{T}_i(x) - y + \mathcal{T}_i(y)\rangle .
		\end{align*}
		Since
		\begin{align*}
		(1+\rho)&\|x-y\|^2+(1+\rho)\|x-\mathcal{T}_i(x) - y + \mathcal{T}_i(y)\|^2 \\&\leq \|x-y\|^2 - \|x-\mathcal{T}_i(x) - y + \mathcal{T}_i(y)\|^2+2(1+\rho)\langle x-y,x-\mathcal{T}_i(x) - y + \mathcal{T}_i(y)\rangle,
		\end{align*}
		we have
		\begin{eqnarray*}
			2(1+\rho)\langle x-y,x-\mathcal{T}_i(x) - y + \mathcal{T}_i(y) \rangle \geq \rho\|x-y\|^2+(2+\rho)\|x-\mathcal{T}_i(x) - y + \mathcal{T}_i(y)\|^2\\
			\langle x-y,\mathcal{T}_i(x)-x +y - \mathcal{T}_i(y) \rangle \leq - \left(\frac{\rho}{2(1+\rho)}\|x-y\|^2+\frac{2+\rho}{2(1+\rho)}\|x-\mathcal{T}_i(x) - y + \mathcal{T}_i(y)\|^2\right).
		\end{eqnarray*}
	\end{proof}

	\section{Analysis of Algorithm~\ref{alg} in Theorem \ref{thh3}}


	The first lemma allows us to find a recursion on the optimality gap for a single step of local method:

\begin{lemma}\label{mainlemma1}	
Under Assumption~\ref{ass:firm} and the condition $ 0 \leq \lambda \leq 1$ we have, for every $k\in\mathbb{N}$,
		\begin{equation}
		\|\hat{x}^{k+1} - x^\star\|^2 \leq\|\hat{x}^{k} - x^\star\|^2+ \lambda(2-\lambda) V_k
		-\frac{1}{2}\lambda(1 - \lambda)\frac{1}{M}\sum^M_{i=1}\big\|g_i(\hat{x}^k)-g_i(x^\star) \big\|^2.
		\end{equation}
	\end{lemma}
	\begin{lemma}\label{lemmaforvar}
		Under Assumption~\ref{ass:firm} and the condition $ 0 \leq \lambda \leq 1$ we have, for every $k\in\mathbb{N}$,
		\begin{equation}
		V_k\leq \lambda^2(H-1)\sum^k_{j=k_p}\frac{3}{M}\sum^M_{i=1}\|x^j_i - \hat{x}^j\|^2
		+
		\sum^k_{j=k_p}\frac{2}{M}\sum^M_{i=1}\|g_i(\hat{x}^j)- g_i(x^\star)\|^2 + 6\sum^k_{j=k_p}\sigma^2.
		\end{equation}
	\end{lemma}


	\subsection{Proof of Lemma~\ref{mainlemma1}}
	Under Assumption~\ref{ass:firm} and under the condition $ 0 \leq \lambda \leq 1$, we have
	\begin{equation}
	\|\hat{x}^{k+1} - x^\star\|^2 \leq\|\hat{x}^{k} - x^\star\|^2+ \lambda(2-\lambda) V_k-\frac{1}{2}\lambda(1 - \lambda)\frac{1}{M}\sum^M_{i=1}\big\|g_i(\hat{x}^k)-g_i(x^\star) \big\|^2.
	\end{equation}
		
	\begin{proof}
		\begin{align*}
		\|\hat{x}^{k+1} - x^\star\|^2 &= \|\hat{x}^{k+1} - \hat{x}^{k} + \hat{x}^{k} - x^\star\|^2 \\& =\|\hat{x}^{k} - x^\star\|^2 + 2\langle\hat{x}^{k+1} - \hat{x}^{k},\hat{x}^{k} - x^\star\rangle + \|\hat{x}^{k+1} - \hat{x}^{k}\|^2\\&
		= \|\hat{x}^{k} - x^\star\|^2 + 2\langle(1-\lambda)\hat{x}^{k} + \lambda\frac{1}{M}\sum^M_{i=1}\mathcal{T}_i(x^k_i) - \hat{x}^{k},\hat{x}^{k} - x^\star\rangle\notag\\
		&\quad+ \|(1-\lambda)\hat{x}^{k} + \lambda\frac{1}{M}\sum^M_{i=1}\mathcal{T}_i(x^k_i) - \hat{x}^{k}\|^2\\&
		= \|\hat{x}^{k} - x^\star\|^2+2\lambda\langle\frac{1}{M}\sum^M_{i=1}\left(\mathcal{T}_i(x_i^k) - \hat{x}^k\right), \hat{x}^k-x^\star\rangle\notag\\
		&\quad+\lambda^2\|\frac{1}{M}\sum^M_{i=1}\left(\mathcal{T}_i(x_i^k) - \hat{x}^k\right)\|^2\\&
		= \|\hat{x}^{k} - x^\star\|^2 + 2\lambda\frac{1}{M}\sum^M_{i=1}\langle\mathcal{T}_i(x_i^k)-x_i^k - \mathcal{T}_i(x^\star) + x^\star, \hat{x}^k - x^\star\rangle\notag\\
		&\quad+\lambda^2\|\frac{1}{M}\sum^M_{i=1}\left(\mathcal{T}_i(x_i^k)-x_i^k - \mathcal{T}_i(x^\star) + x^\star\right)\|^2\\&
		=2\lambda\frac{1}{M}\sum^M_{i=1}\big[\langle\mathcal{T}_i(x_i^k)-x_i^k - \mathcal{T}_i(x^\star) + x^\star, x^k_i - x^\star\rangle\notag \\
		&\quad+ \langle\mathcal{T}_i(x_i^k)-x_i^k - \mathcal{T}_i(x^\star) + x^\star, \hat{x}^k - x^k_i\rangle\big]\\&
		\quad+\|\hat{x}^{k} - x^\star\|^2 +\lambda^2\|\frac{1}{M}\sum^M_{i=1}\left(\mathcal{T}_i(x_i^k)-x_i^k - \mathcal{T}_i(x^\star) + x^\star\right)\|^2\notag\\&
		= \|\hat{x}^{k} - x^\star\|^2 + 2\lambda\frac{1}{M}\sum^M_{i=1}\langle\mathcal{T}_i(x_i^k)-x_i^k - \mathcal{T}_i(x^\star) + x^\star, x^k_i - x^\star\rangle\\&
		\quad+ 2\lambda\frac{1}{M}\sum^M_{i=1} \langle\mathcal{T}_i(x_i^k)-x_i^k - \mathcal{T}_i(x^\star) + x^\star, \hat{x}^k - x^k_i\rangle\notag \\
		&\quad+\lambda^2\big\|\frac{1}{M}\sum^M_{i=1}\left(\mathcal{T}_i(x_i^k)-x_i^k - \mathcal{T}_i(x^\star) + x^\star\right)\big\|^2\notag.
		\end{align*}
		Using Technical Lemma 1,
		\begin{align*}
		\|\hat{x}^{k+1} - x^\star\|^2 &\leq
		\|\hat{x}^{k} - x^\star\|^2 - 2\lambda\frac{1}{M}\sum^M_{i=1}\|\mathcal{T}_i(x_i^k)-x_i^k - \mathcal{T}_i(x^\star) + x^\star \|^2\\&\quad+2\lambda\frac{1}{M}\sum^M_{i=1} \langle\mathcal{T}_i(x_i^k)-x_i^k - \mathcal{T}_i(x^\star) + x^\star, \hat{x}^k - x^k_i\rangle\notag\\
		&\quad+\lambda^2\frac{1}{M}\|\sum^M_{i=1}\left(\mathcal{T}_i(x_i^k)-x_i^k - \mathcal{T}_i(x^\star) + x^\star\right)\|^2\notag
		\end{align*}
		Using the inequality~\eqref{young2}
		\begin{align*}
		\|\hat{x}^{k+1} - x^\star\|^2&\leq \|\hat{x}^{k} - x^\star\|^2 - 2\lambda\frac{1}{M}\sum^M_{i=1}\|\mathcal{T}_i(x_i^k)-x_i^k - \mathcal{T}_i(x^\star) + x^\star \|^2\notag\\
		&\quad+\lambda^2\frac{1}{M}\sum^M_{i=1}\|\mathcal{T}_i(x_i^k)-x_i^k - \mathcal{T}_i(x^\star) + x^\star\|^2\\
		&\quad+2\lambda\frac{1}{M}\sum^M_{i=1}\left(\frac{1}{2}\|\mathcal{T}_i(x_i^k)-x_i^k - \mathcal{T}_i(x^\star) + x^\star\|^2+\frac{1}{2}\|\hat{x}^k - x^k_i\|^2\right)\notag\\&
		= \|\hat{x}^{k} - x^\star\|^2 - \lambda(1 - \lambda)\frac{1}{M}\sum^M_{i=1}\|\mathcal{T}_i(x_i^k)-x_i^k - \mathcal{T}_i(x^\star) + x^\star \|^2\\
		&\quad+\lambda\frac{1}{M}\sum^M_{i=1}\|\hat{x}^k - x^k_i\|^2\\&
		= \|\hat{x}^{k} - x^\star\|^2 - \lambda(1 - \lambda)\frac{1}{M}\sum^M_{i=1}\|\mathcal{T}_i(x_i^k)-x_i^k - \mathcal{T}_i(x^\star) + x^\star \|^2 +\lambda V_k.
		\end{align*}
		Hence,
		\begin{align*}
		\|\hat{x}^{k+1} - x^\star\|^2 &\leq \|\hat{x}^{k} - x^\star\|^2+\lambda V_k\notag\\&
		\quad-\lambda(1 - \lambda)\frac{1}{M}\sum^M_{i=1}\Big\|\mathcal{T}_i(x_i^k)-x_i^k - \mathcal{T}_i(x^\star) + x^\star+\mathcal{T}_i(\hat{x}^k)-\mathcal{T}_i(\hat{x}^k)+\hat{x}^k -\hat{x}^k  \Big\|^2\\&
		=  \|\hat{x}^{k} - x^\star\|^2 +\lambda V_k\notag\\&
		\quad-\lambda(1 - \lambda)\frac{1}{M}\sum^M_{i=1}\Big\|\left(\mathcal{T}_i(x_i^k)-x_i^k -\mathcal{T}_i(\hat{x}^k)+\hat{x}^k\right)+\left(\mathcal{T}_i(\hat{x}^k) -\hat{x}^k - \mathcal{T}_i(x^\star) + x^\star \right) \Big\|^2.
		\end{align*}
		Using the inequality~\eqref{young3}
		\begin{align*}
		\|\hat{x}^{k+1} - x^\star\|^2&\leq\|\hat{x}^{k} - x^\star\|^2 -\frac{1}{2}\lambda(1 - \lambda)\frac{1}{M}\sum^M_{i=1}\Big\|\mathcal{T}_i(\hat{x}^k)-\hat{x}^k - \mathcal{T}_i(x^\star)+x^\star\Big\|^2 \notag\\& \quad+(1-\lambda)\lambda\frac{1}{M}\sum^M_{i=1}\Big\|\mathcal{T}_i(x_i^k) - x_i^k - \mathcal{T}_i(\hat{x}^k)+\hat{x}^k \Big\|^2 + \lambda V_k\\&
		\leq\|\hat{x}^{k} - x^\star\|^2 -\frac{1}{2}\lambda(1 - \lambda)\frac{1}{M}\sum^M_{i=1}\Big\|\mathcal{T}_i(\hat{x}^k)-\hat{x}^k - \mathcal{T}_i(x^\star)+x^\star \Big\|^2  + \lambda(2-\lambda) V_k\\&
		= \|\hat{x}^{k} - x^\star\|^2 -\frac{1}{2}\lambda(1 - \lambda)\frac{1}{M}\sum^M_{i=1}\Big\|\hat{x}^k - \mathcal{T}_i(\hat{x}^k) + \mathcal{T}_i(x^\star)-x^\star \Big\|^2  + \lambda(2-\lambda) V_k.
		\end{align*}
	\end{proof}
	
	\subsection{Proof of Lemma~\ref{lemmaforvar}}
	In this section, we prove the following extended version of Lemma~\ref{lemmaforvar}:
	Under Assumption~\ref{ass:firm} and under the condition $ 0 \leq \lambda \leq 1$, we have
	\begin{align}
	V_k&\leq \lambda^2(H-1)\sum^k_{j=k_p}\frac{3}{M}\sum^M_{i=1}\|x^j_i - \hat{x}^j\|^2\notag\\
	&\quad+
	\sum^k_{j=k_p}\frac{2}{M}\sum^M_{i=1}\|g_i(\hat{x}^j)- g_i(x^\star)\|^2 + 6\sum^k_{j=k_p}\sigma^2.
	\end{align}
	Moreover, for $\lambda \leq \frac{1}{8\max(1,H-1)}$, we have
	\begin{align}
	\label{imp}
	\sum^{k_{p+1}-1}_{k=k_p}&\left(-\frac{1}{2}\lambda(1-\lambda)\frac{1}{M}\sum^M_{i=1}\|\hat{x}^k - \mathcal{T}_i(\hat{x}^k) + \mathcal{T}_i(x^\star)-x^\star\|^2 + \lambda(2-\lambda)V_k\right)\notag\\&	\leq-\frac{\lambda}{3}\sum^{k_{p+1}-1}_{k=k_p}\frac{1}{M}\sum^M_{i=1}\|\hat{x}^k - \mathcal{T}_i(\hat{x}^k) + \mathcal{T}_i(x^\star)-x^\star \|^2+12\lambda^3\sigma^2\sum^{k_{p+1}-1}_{k=k_p}\sigma^2.
	\end{align}
	\begin{proof}
		\begin{align*}
		V_k &= \frac{1}{M} \sum_{i=1}^{M}\left\|x_{i}^{k}-\hat{x}^{k}\right\|^{2} \\&= \frac{1}{M}\sum^M_{i=1}\|x^{k_p}_i - \hat{x}^{k_p} - \lambda\sum^{k}_{j=k_p}g_{i}(x^k_i) - \hat{g}^{j}  \|^2\\&
		= \lambda^2\frac{1}{M}\sum^M_{i=1}\Big\|\sum^k_{j=k_p}(g_i(x^j_i)-\hat{g}^j)\Big\|^2 \\& =\lambda^2\frac{1}{M}\sum^M_{i=1}(k-k_p)\sum^k_{j=k_p}\|g_{i}(x^k_i) -\hat{g}^j\|^2.
		\end{align*}
		Using the property~\eqref{moment},
		\begin{align*}V_k&\leq\lambda^2(H-1)\frac{1}{M}\sum^{M}_{i=1}\sum^k_{j=k_p}\|g_{i}(x^j_i) -\hat{g}^j\|^2\\& \leq\lambda^2(H-1)\frac{1}{M}\sum^{M}_{i=1}\sum^k_{j=k_p}\|g_{i}(x^j_i) \|^2.
		\end{align*}

		Using \eqref{young2}, we have
		\begin{align*}
		\left\|g_{i}(x^k_i)\right\|^{2} &\leq(1+c_1)\left\|g_{i}(x^k_i)-g_i(\hat{x}^k)\right\|^{2}+\left(1+c_1^{-1}\right)\left\|g_i(\hat{x}^k)\right\|^{2}\notag\\&
		\leq (1+c_1)\left\|g_{i}(x^k_i)-g_{i}(\hat{x}^k)\right\|^{2}+\left(1+c_1^{-1}\right)(1+c_2)\left\|g_{i}(\hat{x}^k)-g_{i}(x^\star)\right\|^{2}\\&
		\quad+\left(1+c_1^{-1}\right)\left(1+c_2^{-1}\right)\left\|g_i(x^\star)\right\|^{2}.
		\end{align*}
		Setting $\lambda = 2$ and $\beta = \frac{1}{3}$, we get\\
		\begin{align*}
		&3\left\|g_{i}(x_i^k)-g_{i}(\hat{x}^k)\right\|^{2}+2\left\|g_{i}(\hat{x}^k)-g_{i}(x^\star)\right\|^{2}+6\left\|g_i(x^\star)\right\|^{2}\notag\\
		= {}&3\left\|x^k_i - \mathcal{T}_i(x_i^k)-\hat{x}^k+\mathcal{T}_i(\hat{x}^k)\right\|^{2} + 2\left\|g_{i}(\hat{x}^k)-g_{i}(x^\star)\right\|^{2}+6\left\|g_i(x^\star)\right\|^{2}.
		\end{align*}
		Then
		\begin{equation*}
		\frac{1}{M}\sum^M_{i=1} \|g_i(x_i^k)\|^2 \leq 3\frac{1}{M}\sum^M_{i=1}\|x^k_i - \hat{x}^k\|^2+2\frac{1}{M}\sum^M_{i=1}\|g_i(\hat{x}^k) - g_i(x^\star)\|^2 + 6\sigma^2.
		\end{equation*}
		So, we have
		\begin{equation*}
		V_k\leq \lambda^2(H-1)\sum^k_{j=k_p}\left(3\frac{1}{M}\sum^M_{i=1}\|x^j_i - \hat{x}^j\|^2+2\frac{1}{M}\sum^M_{i=1}\|g_i(\hat{x}^j)- g_i(x^\star)\|^2 + 6\sigma^2\right)
		\end{equation*}
		We get by summation:
		\begin{align*}
		\sum^{k_{p+1}-1}_{k=k_p}V_k &\leq \lambda^2(H-1)\sum^{k_{p+1}-1}_{k=k_p}\sum^k_{j=k_p}\left(3\frac{1}{M}\sum^M_{i=1}\|x^j_i - \hat{x}^j\|^2+2\frac{1}{M}\sum^M_{i=1}\|g_i(\hat{x}^j) - g_i(x^\star)\|^2 + 6\sigma^2\right)\\&
		\leq\lambda^2(H-1)\sum^{k_{p+1}-1}_{k=k_p}\sum^{k_{p+1}-1}_{j=k_p}\left(3\frac{1}{M}\sum^M_{i=1}\|x^j_i - \hat{x}^j\|^2+2\frac{1}{M}\sum^M_{i=1}\|g_i(\hat{x}^j) - g_i(x^\star)\|^2 + 6\sigma^2\right)\\&
		\leq \lambda^2(H-1)^2\sum^{k_{p+1}-1}_{j=k_p}\left(3V_k+2\frac{1}{M}\sum^M_{i=1}\|g_i(\hat{x}^j)- g_i(x^\star)\|^2 + 6\sigma^2\right)
		\end{align*}
		Thus,
		\begin{align*}
		(1-3\lambda^2(H-1)^2)\sum^{k_{p+1}-1}_{k=k_p}V_k&\leq \lambda^2(H-1)^2\sum^{k_{p+1}-1}_{j=k_p}\left(2\frac{1}{M}\sum^M_{i=1}\|g_i(\hat{x}^j) - g_i(x^\star)\|^2 + 6\sigma^2\right)\\
		\sum^{k_{p+1}-1}_{k=k_p}V_k&\leq \frac{\lambda^2(H-1)^2}{(1-3\lambda^2(H-1)^2)}\sum^{k_{p+1}-1}_{j=k_p}\left(2\frac{1}{M}\sum^M_{i=1}\|g_i(\hat{x}^j) - g_i(x^\star)\|^2 + 6\sigma^2\right).
		\end{align*}
		Using $\lambda \leq \frac{1}{8\max(1,H-1)}$, we get
		\begin{equation*}
		\sum^{k_{p+1}-1}_{k=k_p}V_k\leq \frac{16}{15}\lambda^2(H-1)^2\sum^{k_{p+1}-1}_{j=k_p}\left(2\frac{1}{M}\sum^M_{i=1}\|g_i(\hat{x}^j) - g_i(x^\star)\|^2 + 6\sigma^2\right).
		\end{equation*}
		Using this result
		\begin{align*}
		\sum^{k_{p+1}-1}_{k=k_p}&\left(-\frac{1}{2}\lambda(1-\lambda)\frac{1}{M}\sum^M_{i=1}\|\hat{x}^k - \mathcal{T}_i(\hat{x}^k) + \mathcal{T}_i(x^\star)-x^\star\|^2 + \lambda(2-\lambda)V_k\right)\\&
		=-\frac{1}{2}\lambda(1-\lambda)\sum^{k_{p+1}-1}_{k=k_p}\frac{1}{M}\sum^M_{i=1}\|\hat{x}^k - \mathcal{T}_i(\hat{x}^k) + \mathcal{T}_i(x^\star)-x^\star\|^2+\lambda(2-\lambda)\sum^{k_{p+1}-1}_{k=k_p}V_k\\&
		\leq-\frac{1}{2}\lambda(1-\lambda)\sum^{k_{p+1}-1}_{k=k_p}\frac{1}{M}\sum^M_{i=1}\|\hat{x}^k - \mathcal{T}_i(\hat{x}^k) + \mathcal{T}_i(x^\star)-x^\star\|^2\\&\quad+\frac{16}{15}\lambda(2-\lambda)\lambda^2(H-1)^2\sum^{k_{p+1}-1}_{k=k_p}\left(\frac{2}{M}\sum^M_{i=1}\|g_i(\hat{x}^k) - g_i(x^\star)\|^2 + 6\sigma^2\right)\\&
		= -\left(\frac{1}{2}\lambda(1-\lambda)-\frac{16}{15}\lambda(2-\lambda)\lambda^2(H-1)^2\right)\sum^{k_{p+1}-1}_{k=k_p}\frac{1}{M}\sum^M_{i=1}\|\hat{x}^k - \mathcal{T}_i(\hat{x}^k) + \mathcal{T}_i(x^\star)-x^\star\|^2\\&\quad+6\lambda(2-\lambda)\frac{16}{15}\lambda^2(H-1)^2\sum^{k_{p+1}-1}_{k=k_p}\sigma^2\\&
		\leq -\frac{\lambda}{3}\sum^{k_{p+1}-1}_{k=k_p}\frac{1}{M}\sum^M_{i=1}\|\hat{x}^k - \mathcal{T}_i(\hat{x}^k) + \mathcal{T}_i(x^\star)-x^\star\|^2+12\lambda^3(H-1)^2\sum^{k_{p+1}-1}_{k=k_p}\sigma^2.
		\end{align*}
	\end{proof}

	\subsection{Proof of Theorem \ref{thh3}}
	Suppose that $\lambda \leq \frac{1}{8\max(1,H-1)}$ and that  Assumption~\ref{ass:firm} holds. Then, for every $k\in\mathbb{N}$,
	\begin{align}
	\frac{1}{T}\sum^{T-1}_{k=0}\Big\|\hat{x}^k - \mathcal{T}(\hat{x}^k) \Big\|^2 &\leq \frac{3\|\xx - x^\star\|^2}{\lambda T}+36\lambda^2(H-1)^2\sigma^2.
	\end{align}
	\begin{proof}
		Using statment of lemma~\ref{mainlemma1}:
		\begin{equation*}
		\|\hat{x}^{k+1} - x^\star\|^2 \leq \|\hat{x}^{k} - x^\star\|^2 -\frac{1}{2}\lambda(1 - \lambda)\frac{1}{M}\sum^M_{i=1}\Big\|\hat{x}^k - \mathcal{T}_i(\hat{x}^k) + \mathcal{T}_i(x^\star)-x^\star \Big\|^2  + \lambda(2-\lambda) V_k.
		\end{equation*}
		Summing up these inequalities gives
		\begin{align*}
		\sum_{k=0}^{T-1} \|\hat{x}^{k+1} - x^\star\|^2 &\leq \sum^{T-1}_{k=0}|\hat{x}^{k} - x^\star\|^2 \notag\\& +\sum^{T-1}_{k=0}\left(-\frac{1}{2}\lambda(1-\lambda)\frac{1}{M}\sum^M_{i=1}\Big\|\hat{x}^k - \mathcal{T}_i(\hat{x}^k) + \mathcal{T}_i(x^\star)-x^\star \Big\|^2   + \lambda(2-\lambda)V_k\right).
		\end{align*}
		Considering this and using \eqref{imp}
		\begin{align*}
		&\ \ \ \ \sum^{T-1}_{k=0}\left(-\frac{1}{2}\lambda(1-\lambda)\frac{1}{M}\sum^M_{i=1}\Big\|\hat{x}^k - \mathcal{T}_i(\hat{x}^k) + \mathcal{T}_i(x^\star)-x^\star \Big\|^2  + \lambda(2-\lambda)V_k\right) \\& =\sum_{s=1}^p\sum_{j=k_{s-1}}^{k_s-1}\left(-\frac{1}{2}\lambda(1-\lambda)\frac{1}{M}\sum^M_{i=1}\Big\|\hat{x}^k - \mathcal{T}_i(\hat{x}^k) + \mathcal{T}_i(x^\star)-x^\star \Big\|^2  + \lambda(2-\lambda)V_k\right)\\&
		\leq \sum_{s=1}^p \sum_{j=k_p}^{k_p-1}\left(-\frac{1}{2}\lambda(1-\lambda)\frac{1}{M}\sum^M_{i=1}\Big\|\hat{x}^k - \mathcal{T}_i(\hat{x}^k) + \mathcal{T}_i(x^\star)-x^\star \Big\|^2  + \lambda(2-\lambda)V_k\right)\\&
		\leq \sum_{s=1}^p \left( -\frac{\lambda}{3}\sum^{k_{p+1}-1}_{k=k_p}\frac{1}{M}\sum^M_{i=1}\Big\|\hat{x}^k - \mathcal{T}_i(\hat{x}^k) + \mathcal{T}_i(x^\star)-x^\star \Big\|^2 \right)
		+\sum_{s=1}^p\left(12\lambda^3(H-1 )^2\sum^{k_{p+1}-1}_{k=k_p}\sigma^2 \right)\\&
		\leq 12\lambda^3(H-1)^2\sum^{T-1}_{k=0}\sigma^2 - \frac{\lambda}{3}\sum^{T-1}_{k=0}\frac{1}{M}\sum^M_{i=1}\Big\|\hat{x}^k - \mathcal{T}_i(\hat{x}^k) + \mathcal{T}_i(x^\star)-x^\star \Big\|^2.
		\end{align*}
		Hence,
		$$\sum_{k=0}^{T-1} \|\hat{x}^{k+1} - x^\star\|^2 \leq \sum^{T-1}_{k=0}|\hat{x}^{k} - x^\star\|^2+12\lambda^3(H-1)^2\sum^{T-1}_{k=0}\sigma^2-\frac{\lambda}{3}\sum^{T-1}_{k=0}\frac{1}{M}\sum_{i=1}^{M}\Big\|\hat{x}^k - \mathcal{T}_i(\hat{x}^k) + \mathcal{T}_i(x^\star)-x^\star \Big\|^2.$$
		Telescoping this sum:
		\begin{equation*}
		\frac{\lambda}{3}\sum^{T-1}_{k=0}\frac{1}{M}\sum_{i=1}^{M}\Big\|\hat{x}^k - \mathcal{T}_i(\hat{x}^k) + \mathcal{T}_i(x^\star)-x^\star \Big\|^2 \leq
		\|\xx - x^\star\|^2 - \|\hat{x}^{T} - x^\star\|^2 + 12\lambda^3(H-1)^2\sum^{T-1}_{k=0}\sigma^2.
		\end{equation*}
		Using Jensen's inequality \eqref{jensen}:
		\begin{align*}
		\Big\|\hat{x}^k - \frac{1}{M}\sum_{i=1}^{M}\mathcal{T}_i(\hat{x}^k) \Big\|^2 &= \Big\|\frac{1}{M}\sum_{i=1}^{M}\left(\hat{x}^k - \mathcal{T}_i(\hat{x}^k) + \mathcal{T}_i(x^\star)-x^\star \right)\Big\|^2 \\
		&\leq\frac{1}{M}\sum_{i=1}^{M}\Big\|\hat{x}^k - \mathcal{T}_i(\hat{x}^k) + \mathcal{T}_i(x^\star)-x^\star \Big\|^2	.			
		\end{align*}
		Finally, we have
		\begin{align*}
		\frac{\lambda}{3}\sum^{T-1}_{k=0}\frac{1}{M}\sum_{i=1}^M\Big\|\hat{x}^k - \mathcal{T}_i(\hat{x}^k) + \mathcal{T}_i(x^\star)-x^\star \Big\|^2 &\leq \|\xx - x^\star\|^2+12\lambda^3(H-1)^2\sum^{T-1}_{k=0}\sigma^2
		\\
		\frac{\lambda}{3}\sum^{T-1}_{k=0}\Big\|\hat{x}^k - \frac{1}{M}\sum_{i=1}^{M}\mathcal{T}_i(\hat{x}^k) \Big\|^2 &\leq \|\xx - x^\star\|^2+12\lambda^3(H-1)^2T\sigma^2
		\\
		\frac{1}{T}\sum^{T-1}_{k=0}\Big\|\hat{x}^k - \frac{1}{M}\sum_{i=1}^{M}\mathcal{T}_i(\hat{x}^k) \Big\|^2 &\leq \frac{3\|\xx - x^\star\|^2}{\lambda T}+36\lambda^2(H-1)^2\sigma^2
		\\
		\frac{1}{T}\sum^{T-1}_{k=0}\Big\|\hat{x}^k - \mathcal{T}(\hat{x}^k)\Big\|^2 &\leq \frac{3\|\xx - x^\star\|^2}{\lambda T}+36\lambda^2(H-1)^2\sigma^2.
		\end{align*}
	\end{proof}
	\subsection{Proof of Corollary~\ref{cor1}}
	Suppose that $\lambda \leq \frac{1}{8\max(1,H-1)}$ and that  Assumption~\ref{ass:firm} holds. Then a sufficient condition on the number $T$ of iterations to reach $\varepsilon$-accuracy, for any $\varepsilon>0$, is
	\begin{align}
	\frac{T}{H-1} \geq \frac{24\|\hat{x}^0 - x^\star\|^2}{\varepsilon} \max\left\{2, \frac{3 \sigma}{\sqrt{2\varepsilon}} \right\}.\label{corcorr1}
	\end{align}
	\begin{proof}
		\begin{equation*}
		\frac{3\|\xx - x^\star\|^2}{\lambda T}+36\lambda^2(H-1)^2\sigma^2 \leq \varepsilon.
		\end{equation*}
		We have 
		\begin{align*}
		\frac{3\|\hat{x}^0 - x^\star\|^2}{\lambda T}\leq\frac{\varepsilon}{2} &\Rightarrow T\ge\frac{6\|\hat{x}^0 - x^\star\|^2}{\lambda\varepsilon}\\
		36\lambda^2(H-1)^2\sigma^2\leq\frac{\varepsilon}{2}&\Rightarrow\lambda\leq \frac{\sqrt{\varepsilon}}{6\sqrt{2}(H-1)\sigma}.
		\end{align*}
		So, we have
		\begin{equation*}
		\lambda = \min \left\lbrace \frac{1}{8(H-1)},\frac{\sqrt{\varepsilon}}{6\sqrt{2}(H-1)\sigma} \right\rbrace.
		\end{equation*}
		Using this, we get
		\begin{equation*}
		\frac{T}{H-1} \geq \frac{24\|\hat{x}^0 - x^\star\|^2}{\varepsilon} \max\left\{2, \frac{3 \sigma}{\sqrt{2\varepsilon}} \right\}.
		\end{equation*}
	\end{proof}

	\section{Analysis of Algorithm~\ref{alg}: Proof of Theorem \ref{thdis}}
	
	We set $\widetilde{\mathcal{T}}=\frac{1}{M}\sum_{i=1}^M \mathcal{T}_i^H$.
	
	First, we have
\begin{align}
\|x^\dagger-x^\star\|&\leq \|x^\dagger-\widetilde{\mathcal{T}}(x^\star)\|+\|\widetilde{\mathcal{T}}(x^\star)-x^\star\|\\
&= \|\widetilde{\mathcal{T}}(x^\dagger)-\widetilde{\mathcal{T}}(x^\star)\|+\|\widetilde{\mathcal{T}}(x^\star)-x^\star\|\\
&\leq \xi^H \|x^\dagger-x^\star\| + \|\widetilde{\mathcal{T}}(x^\star)-x^\star\|,
\end{align}
so that 
\begin{equation}
\|x^\dagger-x^\star\|\leq \frac{1}{1-\xi^H}\|\widetilde{\mathcal{T}}(x^\star)-x^\star\|.
\end{equation}
Thus, we just have to bound $\|\widetilde{\mathcal{T}}(x^\star)-x^\star\|$:
\begin{align}
\|\widetilde{\mathcal{T}}(x^\star)-x^\star\|&=\|\frac{1}{M}\sum_{i=1}^M \mathcal{T}_i^H (x^\star) -\frac{1}{M}\sum_{i=1}^M \mathcal{T}_i (x^\star)\|\\
&\leq \frac{1}{M}\sum_{i=1}^M \|\mathcal{T}_i^H (x^\star)-\mathcal{T}_i(x^\star)\|\\
&\leq \frac{1}{M}\sum_{i=1}^M \sum_{k=1}^{H-1}\|\mathcal{T}_i^{k+1} (x^\star)-\mathcal{T}_i^k(x^\star)\|\\
&\leq \frac{1}{M}\sum_{i=1}^M \sum_{k=1}^{H-1}\xi^k\|\mathcal{T}_i (x^\star)-x^\star\|\\
&= \frac{1}{M}\xi\frac{1-\xi^{H-1}}{1-\xi}\sum_{i=1}^M \|\mathcal{T}_i (x^\star)-x^\star\|
\end{align}
Hence, 
\begin{align}
\|x^\dagger-x^\star\|&\leq S,
\end{align}
where
\begin{align}
S&= \frac{\xi}{1-\xi}\frac{1-\xi^{H-1}}{1-\xi^H}\frac{1}{M}\sum_{i=1}^M \|\mathcal{T}_i (x^\star)-x^\star\|
\end{align}
 
	\section{Analysis of Algorithm~\ref{alg2}}
	
	We first derive two lemmas, which will be combined to prove Theorem \ref{therrm2}.
	
	The first lemma provides a recurrence property, for one iteration of Algorithm 2:

	\begin{lemma} \label{mainlemma2}
		Under Assumption \ref{ass:s2}, for every $k\in\mathbb{N}$,
		\begin{equation*}
		\|\hat{x}^{k+1} - x^\star\|^2\leq  \left(1-\frac{\lambda\rho}{1+\rho}\right)\|\hat{x}^{k} - x^\star\|^2  + \frac{5}{2}\lambda V_k-\frac{1}{2}\lambda\left( \frac{1}{2}- \lambda\right)\frac{1}{M}\sum^M_{i=1}\big\|g_i(x_i^k)-g_i(x^\star) \big\|^2.
		\end{equation*}
	\end{lemma}

	We now bound the variance $V_k$ for one iteration, using the contraction property:
	

	\begin{lemma}\label{lemmavar2}
		Under Assumption \ref{ass:s2} and if $\lambda<\frac{p}{15}$ we have, for every $k\in\mathbb{N}$,
		\begin{equation*}
		V_k \leq \frac{2}{p}\left(1-\frac{p}{4}+\frac{5}{p}\lambda^2\right)V_k+60\frac{\lambda^2}{p^2}\sigma^2-\frac{2}{p}\mathbb{E}[V_{k+1}]+20\frac{\lambda^2}{p^2}\frac{1}{M}\sum^M_{i=1}\left\|g_i(\hat{x}^k)-g_i(x^\star)\right\|^2.
		\end{equation*}
	\end{lemma}

	\subsection{Proof of Lemma~\ref{mainlemma2}} 
	Under Assumption \ref{ass:s2}, for every $k\in\mathbb{N}$ and $ 0 \leq \lambda \leq 1$, we have
	\begin{align}
	\|&\hat{x}^{k+1} - x^\star\|^2\leq  \left(1-\frac{\lambda\rho}{1+\rho}\right)\|\hat{x}^{k} - x^\star\|^2  + \frac{5}{2}\lambda V_k-\frac{1}{2}\lambda\left( \frac{1}{2}- \lambda\right)\frac{1}{M}\sum^M_{i=1}\big\|g_i(x_i^k)-g_i(x^\star) \big\|^2.\label{eqlcor}
	\end{align}

	\begin{proof}
		\begin{align*}
		x_i^{k+1} &= (1-\lambda)x_i^k+\lambda\mathcal{T}_i(x_i^k)\\
		&=x_i^k - \lambda\left(x_i^k - \mathcal{T}_i(x_i^k)\right) \\
		&= x_i^k - \lambda g_i(x_i^k).
		\end{align*}
		So, we have
		\begin{align*}
		\|\hat{x}^{k+1} - x^\star\|^2 &= \|\hat{x}^{k+1} - \hat{x}^{k} + \hat{x}^{k} - x^\star\|^2\\& = \|\hat{x}^{k} - x^\star\|^2 + 2\langle\hat{x}^{k+1} - \hat{x}^{k},\hat{x}^{k} - x^\star\rangle + \|\hat{x}^{k+1} - \hat{x}^{k}\|^2\\&
		= \|\hat{x}^{k} - x^\star\|^2 + 2\langle(1-\lambda)\hat{x}^{k} + \lambda\frac{1}{M}\sum^M_{i=1}\mathcal{T}_i(x^k_i) - \hat{x}^{k},\hat{x}^{k} - x^\star\rangle\notag\\&
		\quad+ \Big\|(1-\lambda)\hat{x}^{k} + \lambda\frac{1}{M}\sum^M_{i=1}\mathcal{T}_i(x^k_i) - \hat{x}^{k}\Big\|^2\\&
		= \|\hat{x}^{k} - x^\star\|^2+2\lambda\langle\frac{1}{M}\sum^M_{i=1}\mathcal{T}_i(x_i^k) - \hat{x}^k, \hat{x}^k-x^\star\rangle\notag\\&
		\quad+\lambda^2\Big\|\frac{1}{M}\sum^M_{i=1}\left(\mathcal{T}_i(x_i^k) - \hat{x}^k\right)\Big\|^2\\&
	=	\|\hat{x}^{k} - x^\star\|^2 + 2\lambda\frac{1}{M}\sum^M_{i=1}\langle\mathcal{T}_i(x_i^k)-x_i^k - \mathcal{T}_i(x^\star) + x^\star, \hat{x}^k - x^\star\rangle\notag\\&
		\quad+\lambda^2\Big\|\frac{1}{M}\sum^M_{i=1}\left(\mathcal{T}_i(x_i^k)-x_i^k - \mathcal{T}_i(x^\star) + x^\star\right)\Big\|^2\\&
		=\|\hat{x}^{k} - x^\star\|^2+\lambda^2\Big\|\frac{1}{M}\sum^M_{i=1}\left(\mathcal{T}_i(x_i^k)-x_i^k - \mathcal{T}_i(x^\star) + x^\star\right)\Big\|^2\\& 
		\quad+ 2\lambda\frac{1}{M}\sum^M_{i=1}\langle\mathcal{T}_i(x_i^k)-x_i^k - \mathcal{T}_i(x^\star) + x^\star, x^k_i - x^\star\rangle \\
		&\quad+ \langle\mathcal{T}_i(x_i^k)-x_i^k - \mathcal{T}_i(x^\star) + x^\star, \hat{x}^k - x^k_i\rangle\notag\\&
		=  2\lambda\frac{1}{M}\sum^M_{i=1} \langle\mathcal{T}_i(x_i^k)-x_i^k - \mathcal{T}_i(x^\star) + x^\star, \hat{x}^k - x^k_i\rangle+\|\hat{x}^{k} - x^\star\|^2\notag\\&
		\quad+\frac{2\lambda}{M}\sum^M_{i=1}\langle\mathcal{T}_i(x_i^k)-x_i^k - \mathcal{T}_i(x^\star) + x^\star, x^k_i - x^\star\rangle\\
		&\quad+\lambda^2\Big\|\frac{1}{M}\sum^M_{i=1}\left(\mathcal{T}_i(x_i^k)-x_i^k - \mathcal{T}_i(x^\star) + x^\star\right)\Big\|^2.\notag
		\end{align*}
		Using Technical Lemma 2, 
		\begin{align*}
		\|\hat{x}^{k+1} &- x^\star\|^2 \leq \|\hat{x}^{k} - x^\star\|^2 \notag\\
		&- 2\lambda\frac{1}{M}\sum_{i=1}^{M}\left(\frac{\rho}{2(1+\rho)}\|x^k_i-x^\star\|^2\right) \notag\\
		&- 2\lambda\frac{1}{M}\sum_{i=1}^{M}\frac{2+\rho}{2(1+\rho)}\Big\|\mathcal{T}_i(x_i^k) - x_i^k - \mathcal{T}_i(x^\star) + x^\star\Big\|^2\\&
		+ 2\lambda\frac{1}{M}\sum^M_{i=1} \langle\mathcal{T}_i(x_i^k)-x_i^k - \mathcal{T}_i(x^\star) + x^\star, \hat{x}^k - x^k_i\rangle\notag \\
		&+\lambda^2\Big\|\frac{1}{M}\sum^M_{i=1}\left(\mathcal{T}_i(x_i^k)-x_i^k - \mathcal{T}_i(x^\star) + x^\star\right)\Big\|^2.\notag
		\end{align*}
		\begin{align*}
		&\|\hat{x}^{k+1} - x^\star\|^2  \leq\|\hat{x}^{k} - x^\star\|^2 - \frac{\lambda\rho}{(1+\rho)}\Big\|\frac{1}{M}\sum_{i=1}^{M}(x_i^k-x^\star)\Big\|^2\\
		&-2\lambda\frac{2+\rho}{2(1+\rho)}\frac{1}{M}\sum_{i=1}^{M}\Big\|\mathcal{T}_i(x_i^k) - x_i^k-\mathcal{T}_i(x^\star)+x^\star\Big\|^2 \notag\\&
		+ 2\lambda\frac{1}{M}\sum^M_{i=1} \langle\mathcal{T}_i(x_i^k)-x_i^k - \mathcal{T}_i(x^\star) + x^\star, \hat{x}^k - x^k_i\rangle +\lambda^2\frac{1}{M}\sum^M_{i=1}\Big\|\left(\mathcal{T}_i(x_i^k)-x_i^k - \mathcal{T}_i(x^\star) + x^\star\right)\Big\|^2.
		\end{align*}
		Using the inequality~\eqref{young2}
		\begin{align*}
		\|\hat{x}^{k+1} - x^\star\|^2 &\leq \|\hat{x}^{k} - x^\star\|^2\left(1-\frac{\lambda\rho}{1+\rho}\right)+ 2\lambda\frac{1}{M}\sum^M_{i=1} \left(\frac{1}{4}\Big\|\mathcal{T}_i(x_i^k)-x_i^k - \mathcal{T}_i(x^\star) + x^\star\Big\|^2\right.\\
	&\quad\left.+ \Big\|\hat{x}^k - x^k_i\Big\|^2\right)
		-2\lambda\frac{2+\rho}{2(1+\rho)}\frac{1}{M}\sum_{i=1}^{M}\Big\|\mathcal{T}_i(x_i^k) - x_i^k-\mathcal{T}_i(x^\star)+x^\star\Big\|^2\\
		&\quad+\lambda^2\frac{1}{M}\sum^M_{i=1}\Big\|\left(\mathcal{T}_i(x_i^k)-x_i^k - \mathcal{T}_i(x^\star) + x^\star\right)\Big\|^2\\&
		\leq\|\hat{x}^{k} - x^\star\|^2\left(1-\frac{\lambda\rho}{1+\rho}\right) + \left[\lambda^2+\frac{1}{2}\lambda-\frac{\lambda(2+\rho)}{1+\rho}\right]\\
		&\quad\times\frac{1}{M}\sum_{i=1}^{M}\Big\|\left(\mathcal{T}_i(x_i^k)-x_i^k - \mathcal{T}_i(x^\star) + x^\star\right)\Big\|^2+2\lambda V_k.
		\end{align*}
		Hence,
		\begin{align*}
		\|\hat{x}^{k+1} - x^\star\|^2 &\leq\left(1-\frac{\lambda\rho}{1+\rho}\right)\|\hat{x}^{k} - x^\star\|^2+2\lambda V_k \notag\\
		&\quad-\lambda\left(\frac{1}{2} - \lambda\right   )\frac{1}{M}\sum^M_{i=1}\Big\|\mathcal{T}_i(x_i^k)-x_i^k - \mathcal{T}_i(x^\star) + x^\star+\mathcal{T}_i(\hat{x}^k)-\mathcal{T}_i(\hat{x}^k)+\hat{x}^k -\hat{x}^k  \Big\|^2 \\&
		= \left(1-\frac{\lambda\rho}{1+\rho}\right)\|\hat{x}^{k} - x^\star\|^2 +2\lambda V_k\notag\\&
		\quad-\lambda\left(\frac{1}{2} - \lambda\right   )\frac{1}{M}\sum^M_{i=1}\Big\|\left(\mathcal{T}_i(x_i^k)-x_i^k -\mathcal{T}_i(\hat{x}^k)+\hat{x}^k\right)\\
		&\quad+\left(\mathcal{T}_i(\hat{x}^k) -\hat{x}^k - \mathcal{T}_i(x^\star) + x^\star \right) \Big\|^2.
		\end{align*}
		Using~\eqref{young3}, we have
		\begin{align*}
		\|\hat{x}^{k+1} - x^\star\|^2&\leq\left(1-\frac{\lambda\rho}{1+\rho}\right)\|\hat{x}^{k} - x^\star\|^2 -\frac{1}{2}\lambda\left(\frac{1}{2} - \lambda\right   )\frac{1}{M}\sum^M_{i=1}\Big\|\mathcal{T}_i(\hat{x}^k)-\hat{x}^k - \mathcal{T}_i(x^\star)+x^\star\Big\|^2\notag\\&
		+ \lambda\left(\frac{1}{2} - \lambda\right   )\frac{1}{M}\sum^M_{i=1}\Big\|\mathcal{T}_i(x_i^k) - x_i^k - \mathcal{T}_i(\hat{x}^k)+\hat{x}^k \Big\|^2 + 2\lambda V_k\\&
		\leq\left(1-\frac{\lambda\rho}{1+\rho}\right)\|\hat{x}^{k} - x^\star\|^2 + \lambda\left(2+\frac{1}{2}-\lambda\right) V_k\notag\\&
		-\frac{1}{2}\lambda\left( \frac{1}{2}- \lambda\right)\frac{1}{M}\sum^M_{i=1}\Big\|\mathcal{T}_i(\hat{x}^k)-\hat{x}^k - \mathcal{T}_i(x^\star)+x^\star \Big\|^2.
		\end{align*}
		Finally, we have
		\begin{equation}
		\|\hat{x}^{k+1} - x^\star\|^2		\leq\left(1-\frac{\lambda\rho}{1+\rho}\right)\|\hat{x}^{k} - x^\star\|^2 -\frac{1}{2}\lambda\left( \frac{1}{2}- \lambda\right)\frac{1}{M}\sum^M_{i=1}\Big\|\hat{x}^k - \mathcal{T}_i(\hat{x}^k) + \mathcal{T}_i(x^\star)-x^\star \Big\|^2+ \frac{5}{2}\lambda V_k.
		\end{equation}
	\end{proof}
	\subsection{Proof of Lemma~\ref{lemmavar2}}
	Under Assumption \ref{ass:s2} and if $\lambda<\frac{p}{15}$, we have, for every $k\in\mathbb{N}$,
	\begin{equation}
	V_k \leq \frac{2}{p}\left(1-\frac{p}{4}+\frac{5}{p}\lambda^2\right)V_k+20\frac{\lambda^2}{p^2}\frac{1}{M}\sum^M_{i=1}\left\|g_i(\hat{x}^k)-g_i(x^\star)\right\|^2+60\frac{\lambda^2}{p^2}\sigma^2-\frac{2}{p}\mathbb{E}[V_{k+1}].
	\end{equation}
	\begin{proof}
		If communication happens, $V_k = 0$. Therefore,
		\begin{align*}
		\mathbb{E}[V_{k+1}] &= (1-p)\frac{1}{M}\sum^M_{i=1}\left\|\hat{x}^k-\lambda\hat{g}^k-x_i^k+\lambda g_i(x_i^k)\right\|^2
		=(1-p)\frac{1}{M}\sum^M_i\left\|\hat{x}^k-x_i^k\right\|^2\notag\\&+(1-p)\lambda^2\frac{1}{M}\sum^M_{i=1}\left\|g_i(x_i^k)-\hat{g}^k\right\|^2+2(1-p)\lambda\frac{1}{M}\sum^M_{i=1}\langle \hat{x}^k - x^k_i, g_i(x_i^k) - \hat{g}^k\rangle
		\end{align*}
		Using Young's inequality~\eqref{young2},
		\begin{align*}
		\mathbb{E}[V_{k+1}]&\leq(1-p)\frac{1}{M}\sum^M_i\left\|\hat{x}^k-x_i^k\right\|^2+(1-p)\lambda^2\frac{1}{M}\sum^M_{i=1}\left\|g_i(x_i^k)-\hat{g}^k\right\|^2\notag\\&
		\quad+\frac{p}{4}(1-p)\frac{1}{M}\sum^M_i\left\|\hat{x}^k-x_i^k\right\|^2+(1-p)\frac{4}{p}\lambda^2\frac{1}{M}\sum^M_{i=1}\left\|g_i(x_i^k)-\hat{g}^k\right\|^2.
		\end{align*}
		Using our notations, 
		\begin{align*}
		\frac{p}{2}V_k &\leq \left(1-\frac{p}{2}\right)V_k + (1-p)\left( \lambda^2 + \frac{4}{p}\lambda^2\right)\frac{1}{M}\sum^M_{i=1}\left\|g_i(x_i^k)-\hat{g}^k\right\|^2 - \mathbb{E}[V_{k+1}]+(1-p)\frac{p}{4}V_k\\&
		\leq \left(1-\frac{p}{2}+(1-p)\frac{p}{4}\right)V_k+(1-p)\lambda^2\left(1+\frac{4}{p}\right)\frac{1}{M}\sum^M_{i=1}\left\|g_i(x_i^k)-\hat{g}^k\right\|^2 - \mathbb{E}[V_{k+1}]\\&
		\leq \left(1-\frac{p}{2}+(1-p)\frac{p}{4}\right)V_k+\frac{5}{p}(1-p)\lambda^2\frac{1}{M}\sum^M_{i=1}\left\|g_i(x_i^k)\right\|^2 - \mathbb{E}[V_{k+1}].
		\end{align*}
		Applying the same technique, we get:
		\begin{align*}
		\left\|g_{i}(x^k_i)\right\|^{2} &\leq(1+c_1)\left\|g_{i}(x^k_i)-g_i(\hat{x}^k)\right\|^{2}+\left(1+c_1^{-1}\right)\left\|g_i(\hat{x}^k)\right\|^{2}\notag\\
		&\leq (1+c_1)\left\|g_{i}(x^k_i)-g_{i}(\hat{x}^k)\right\|^{2}+\left(1+c_1^{-1}\right)(1+c_2)\left\|g_{i}(\hat{x}^k)-g_{i}(x^\star)\right\|^{2}\\
		&\quad+\left(1+c_1^{-1}\right)\left(1+c_2^{-1}\right)\left\|g_i(x^\star)\right\|^{2}.\notag
		\end{align*}
		Setting $c_1 = 2$, $c_2 = \frac{1}{3}$, we get
		\begin{align*}
		3\left\|g_{i}(x_i^k)-g_{i}(\hat{x}^k)\right\|^{2}&+2\left\|g_{i}(\hat{x}^k)-g_{i}(x^\star)\right\|^{2}+6\left\|g_i(x^\star)\right\|^{2}\\& 
		= 3\left\|x^k_i - \mathcal{T}_i(x_i^k)-\hat{x}^k+\mathcal{T}_i(\hat{x}^k)\right\|^{2} + 2\left\|g_{i}(\hat{x}^k)-g_{i}(x^\star)\right\|^{2}+6\left\|g_i(x^\star)\right\|^{2}.	\notag	\end{align*}
		By averaging,
		\begin{equation*}
		\frac{1}{M}\sum^M_{i=1} \|g_i(x_i^k)\|^2 \leq 3\frac{1}{M}\sum^M_{i=1}\|x^k_i - \hat{x}^k\|^2+2\frac{1}{M}\sum^M_{i=1}\|g_i(\hat{x}^k) - g_i(x^\star)\|^2 + 6\sigma^2.
		\end{equation*}
		Using this inequality,
		\begin{align*}
		\frac{p}{2}V_k &\leq \left(1-\frac{p}{2}+(1-p)\frac{p}{4}\right)V_k+\frac{5}{p}(1-p)\lambda^2\left(3V_k+2\frac{1}{M}\sum^M_{i=1}\left\|g_i(\hat{x}^k)-g_i(x^\star)\right\|^2+6\sigma^2\right)\\
		&\quad-\mathbb{E}[V_{k+1}]\\&
		\leq \left(1-\frac{p}{2}+(1-p)\left(\frac{p}{4}+\frac{5}{p}\lambda^2\right)\right)V_k\\
		&\quad+\frac{5}{p}(1-p)\lambda^2\left(2\frac{1}{M}\sum^M_{i=1}\left\|g_i(\hat{x}^k)-g_i(x^\star)\right\|^2+6\sigma^2\right)-\mathbb{E}[V_{k+1}]
		\\& \leq \left(1-\frac{p}{4}+\frac{5}{p}\lambda^2\right)V_k+\frac{10}{p}\lambda^2\frac{1}{M}\sum^M_{i=1}\left\|g_i(\hat{x}^k)-g_i(x^\star)\right\|^2+30\frac{\lambda^2}{p}\sigma^2-\mathbb{E}[V_{k+1}].
		\end{align*}
		Finally, we get
		\begin{equation*}
		V_k \leq \frac{2}{p}\left(1-\frac{p}{4}+\frac{5}{p}\lambda^2\right)V_k+20\frac{\lambda^2}{p^2}\frac{1}{M}\sum^M_{i=1}\left\|g_i(\hat{x}^k)-g_i(x^\star)\right\|^2+60\frac{\lambda^2}{p^2}\sigma^2-\frac{2}{p}\mathbb{E}[V_{k+1}].
		\end{equation*}
		
	\end{proof}
	\subsection{Proof of Theorem~\ref{therrm2}}
	For every $k\in\mathbb{N}$, let $\Psi^k$ be the Lyapunov function defined as:
	
	\begin{equation}
	\Psi^k \coloneqq     	\|\hat{x}^{k} - x^\star\|^2  + \frac{5\lambda}{p}V_{k}.
	\end{equation}
	
	Under Assumption \ref{ass:s2} and if $\lambda<\frac{p}{15}$, we have, for every $k\in\mathbb{N}$,
	\begin{align}
	\mathbb{E}\Psi^{k}\leq \left(1-\min\left(\frac{\lambda\rho}{1+\rho},\frac{p}{5}\right)\right)^k\Psi^0+\frac{150}{\min\left(\frac{\lambda\rho}{1+\rho},\frac{p}{5}\right)p^2}\lambda^3\sigma^2.
	\end{align}
	\begin{proof}
		Using Lemma~\ref{mainlemma2},
		\begin{equation*}
		\|\hat{x}^{k+1} - x^\star\|^2\leq  \left(1-\frac{\lambda\rho}{1+\rho}\right)\|\hat{x}^{k} - x^\star\|^2 -\frac{1}{2}\lambda\left( \frac{1}{2}- \lambda\right)\frac{1}{M}\sum^M_{i=1}\Big\|g_i(\hat{x}^k)-g_i(x^\star) \Big\|^2  + \frac{5}{2}\lambda V_k.
		\end{equation*}	
		Using Lemma~\ref{lemmavar2},
		\begin{align*}
		\|\hat{x}^{k+1} - x^\star\|^2&\leq \left(1-\frac{\lambda\rho}{1+\rho}\right)\|\hat{x}^{k} - x^\star\|^2 -\frac{1}{2}\lambda\left( \frac{1}{2}- \lambda\right)\frac{1}{M}\sum^M_{i=1}\Big\|g_i(\hat{x}^k)-g_i(x^\star) \Big\|^2\\&
		\quad+\frac{5}{2}\lambda\frac{2}{p}\left(\left(1-\frac{p}{4}+\frac{5}{p}\lambda^2\right)V_k-\mathbb{E}[V_{k+1}]\right)+\frac{5}{2}\lambda\frac{20}{p^2}\lambda^2\frac{1}{M}\sum^M_{i=1}\Big\|g_i(\hat{x}^k)-g_i(x^\star) \Big\|^2\notag\\
		&\quad+60\frac{\lambda^2}{p^2}\frac{5}{2}\lambda\sigma^2\notag\\&
		\leq \left(1-\frac{\lambda\rho}{1+\rho}\right)\|\hat{x}^{k} - x^\star\|^2 + \lambda\left(\frac{50}{p^2}\lambda^2-\frac{1}{2}\left( \frac{1}{2}- \lambda\right)\right)\frac{1}{M}\sum^M_{i=1}\Big\|g_i(\hat{x}^k)-g_i(x^\star) \Big\|^2 \notag\\&
		\quad+\frac{5}{2}\lambda\frac{2}{p}\left(\left(1-\frac{p}{4}+\frac{5}{p}\lambda^2\right)V_k-\mathbb{E}[V_{k+1}]\right)+150\frac{\lambda^3}{p^2}\sigma^2.
		\end{align*}
		If $\lambda\leq \frac{p}{15}$, we have
		\begin{equation*}
		\|\hat{x}^{k+1} - x^\star\|^2  \leq
		\left(1-\frac{\lambda\rho}{1+\rho}\right)\|\hat{x}^{k} - x^\star\|^2
		+\frac{5\lambda}{p}\left(\left(1-\frac{p}{4}+\frac{5}{p}\lambda^2\right)V_k-\mathbb{E}[V_{k+1}]\right)+150\frac{\lambda^3}{p^2}\sigma^2.				\end{equation*}
		We have the contraction property
		\begin{equation*}
		\|\hat{x}^{k+1} - x^\star\|^2  + \frac{5\lambda}{p}\mathbb{E}[V_{k+1}]\leq
		\left(1-\frac{\lambda\rho}{1+\rho}\right)\|\hat{x}^{k} - x^\star\|^2 +\frac{5\lambda}{p}\left(1-\frac{p}{5}\right)V_k+150\frac{\lambda^3}{p^2}\sigma^2.
		\end{equation*}
		Define the Lyapunov function:
		\begin{equation*}
		\Psi^k =     	\|\hat{x}^{k} - x^\star\|^2  + \frac{5\lambda}{p}V_{k}.
		\end{equation*}
		Using the law of total expectation, we get
		\begin{equation*}
		\mathbb{E}\Psi^{k+1} \leq \left(1-\min\left(\frac{\lambda\rho}{1+\rho},\frac{p}{5}\right)\right)\Psi^k+150\frac{\lambda^3}{p^2}\sigma^2.
		\end{equation*}
		Finally we get
		\begin{equation*}
		\mathbb{E}\Psi^{T}\leq \left(1-\min\left(\frac{\lambda\rho}{1+\rho},\frac{p}{5}\right)\right)^T\Psi^0+\frac{150}{\min\left(\frac{\lambda\rho}{1+\rho},\frac{p}{5}\right)}\frac{\lambda^3}{p^2}\sigma^2.
		\end{equation*}
	\end{proof}
	\subsection{Proof of Corollary~\ref{cor2}}
	Under Assumption \ref{ass:s2} and if $\lambda<\frac{p}{15}$, for any $\varepsilon>0$, $\varepsilon$-accuracy  is reached after $T$ iterations, with
	\begin{align}
	T \geq \max \left\{\frac{15(1 + \rho)}{\rho p}, \frac{18 \sigma (1 + \rho)^\frac13}{p \rho^\frac32 \varepsilon^\frac12}, \frac{40 \sigma^\frac23(1 + \rho)}{p \rho \varepsilon^\frac13} \right\}\log \frac{2 \Psi_0}{\varepsilon}.
	\end{align}
	\begin{proof}
		We start from 
		\begin{equation*}
		\left[1 - \min\left\{\frac{\lambda \rho}{\rho + 1}, \frac{p}{5} \right\} \right]^k \Psi_0 + \frac{150 \lambda^3 \sigma^2}{p^2 \min\left\{\frac{\lambda \rho}{\rho + 1}, \frac{p}{5} \right\}} \leq \varepsilon.
		\end{equation*}
		Regarding the second term, if 
		$\displaystyle
		150 \lambda^3 \sigma^2 \leq \frac12 p^2 \varepsilon \min\left\{\frac{\lambda \rho}{\rho + 1}, \frac{p}{5} \right\}$, then
		$\displaystyle
		\begin{cases}
		150 \lambda^3 \sigma^2 \leq \frac12 p^2 \varepsilon \frac{\lambda \rho}{\rho + 1},\\
		150 \lambda^3 \sigma^2 \leq \frac{ p^3 \varepsilon}{10}
		\end{cases}$, 
		
		\noindent so that 
		$\lambda \leq \min \left\{\frac{p}{18\sigma}\sqrt{\frac{\varepsilon \rho}{\rho + 1}}, \frac{p \varepsilon^{\frac13}}{40 \sigma^{\frac23}} \right\}.
		$
		
		Regarding the first term, and using the fact that $\lambda<\frac{p}{15}$,
		if $\displaystyle
		\left[1 - \min\left\{\frac{\lambda \rho}{\rho + 1}, \frac{p}{5} \right\} \right]^T \Psi_0 \leq \frac{\varepsilon}{2}$,
		
		\noindent then $T \geq \max \left\{\frac{1 + \rho}{\lambda \rho}, \frac{5}{p}\right\} \log \frac{2 \Psi_0}{\varepsilon}$, so that 
		$\lambda = \min \left\{\frac{p}{15}, \frac{p}{18\sigma}\sqrt{\frac{\varepsilon \rho}{\rho + 1}}, \frac{p \varepsilon^{\frac13}}{40 \sigma^{\frac23}} \right\}.
		$
		
		Finally, we get
		\begin{align*}
		T &\geq \max \left\{\frac{5}{p}, \frac{15(1 + \rho)}{\rho p}, \frac{18 \sigma (1 + \rho)^\frac13}{p \rho^\frac32 \varepsilon^\frac12}, \frac{40 \sigma^\frac23(1 + \rho)}{p \rho \varepsilon^\frac13}  \right\}\log \frac{2 \Psi_0}{\varepsilon} \\
		&= \max \left\{\frac{15(1 + \rho)}{\rho p}, \frac{18 \sigma (1 + \rho)^\frac13}{p \rho^\frac32 \varepsilon^\frac12}, \frac{40 \sigma^\frac23(1 + \rho)}{p \rho \varepsilon^\frac13}  \right\}\log \frac{2 \Psi_0}{\varepsilon}.
		\end{align*}
	\end{proof}


\end{document}